\documentclass[10pt,twocolumn,letterpaper]{article}

\usepackage{iccv}
\usepackage{times}
\usepackage{epsfig}
\usepackage{graphicx}
\usepackage{amsmath}
\usepackage{amsthm}
\usepackage{amssymb}
\usepackage{subfigure}
\usepackage{dsfont}
\usepackage{algorithmic}
\usepackage{float}
\usepackage{color}

\usepackage{soul}
\usepackage{wrapfig}
\usepackage{algorithm,algorithmic}

\graphicspath{{images/}}

\newcommand{\comment}[1]{}
\newcommand{\raphaelrmk}[1]{{\color{blue}{\bf #1}}}

\newcommand{\bx}{\mathbf{x}}

\DeclareMathOperator*{\argmax}{arg\,max}

\newcommand{\RS}{{\bf Rs}}
\newcommand{\FU}{{\bf FUs}}
\newcommand{\CU}{{\bf CUs}}
\newcommand{\PFU}{{\bf pFUs}}
\newcommand{\PCU}{{\bf pCUs}}

\iccvfinalcopy 

\def\iccvPaperID{2080} 

\ificcvfinal\fi
\begin{document}


\title{Introducing Geometry in Active Learning for Image Segmentation}

\author{Ksenia Konyushkova\\
EPFL\\
{\tt\small ksenia.konyushkova@epfl.ch}
\and
Raphael Sznitman\\
University of Bern\\
{\tt\small raphael.sznitman@artorg.unibe.ch}
\and
Pascal Fua\\
EPFL\\
{\tt\small pascal.fua@epfl.ch}
}

\maketitle


\begin{abstract}

We propose  an Active  Learning approach to  training a  segmentation classifier
that exploits geometric priors to streamline  the annotation process in 3D image
volumes.  To  this end, we use  these priors not  only to select voxels  most in
need of  annotation but  to guarantee that  they lie on  2D planar  patch, which
makes it much easier  to annotate than if they were  randomly distributed in the
volume. A simplified version of this approach is effective in natural 2D images.

We evaluated  our approach on  Electron Microscopy and Magnetic  Resonance image
volumes, as well  as on natural images.  Comparing our  approach against several
accepted baselines demonstrates a marked performance increase.

\end{abstract}


\section{Introduction}
\label{sec:introduction}

Machine  Learning  techniques  are  a  key component  of  modern  approaches  to
segmentation,  making the need for sufficient   amounts  of  training  data critical. 
As far as  images of everyday scenes  are concerned, this is  addressed by
compiling   ever   larger   training   databases  and   obtaining   the   ground
truth via  crowd-sourcing~\cite{Long13,Lin14}.   By  contrast,  in  specialized
domains such as  biomedical image processing, this is not always  an option both
because the images can only be acquired using very sophisticated instruments and
because  only experts  whose  time  is scarce  and  precious  can annotate  them
reliably.

Active Learning (AL) is an established way to reduce this labeling  workload by  automatically deciding  which parts  of the  image an annotator should label to train the system as quickly  as possible and with minimal amounts of manual intervention.  
However, most AL  techniques used in Computer Vision, such
as~\cite{Kapoor07,Joshi09,Vezhnevets12,Maiora12},   are   inspired  by   earlier
methods developed  primarily   for general tasks  or   Natural Language
Processing~\cite{Tong02,Lewis94}. As such, they  rarely   account  for   the  specific
difficulties or exploit the opportunities  that arise when annotating individual
pixels in 2D images and 3D voxels in image volumes.

\begin{figure}[t]
 \begin{center}
\begin{tabular}{cc}
  \hspace{-0.3cm}\includegraphics[height=0.4\linewidth]{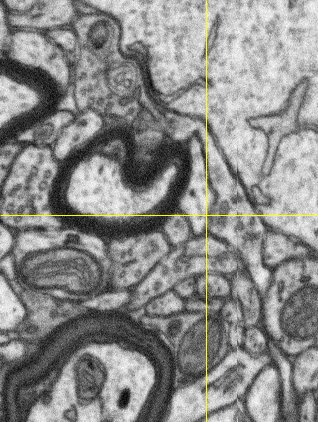}&
  \hspace{-0.3cm}\includegraphics[height=0.4\linewidth]{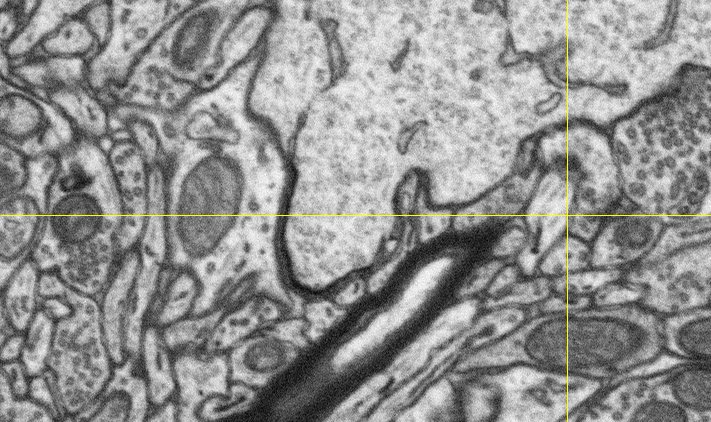}\\[-0.1cm]
  (yz)&(xy)\\
  \hspace{-0.3cm}\includegraphics[width=0.302\linewidth]{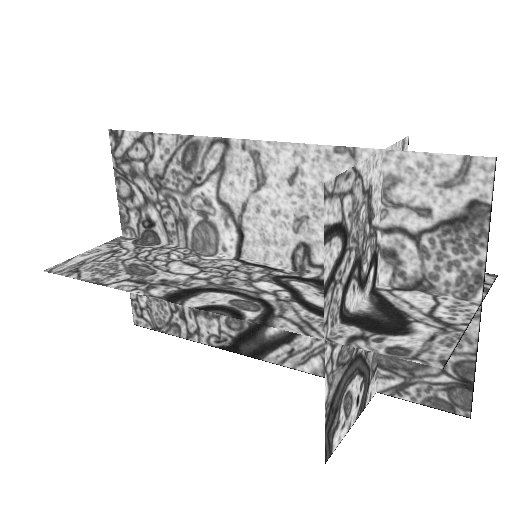}&
  \hspace{-0.3cm}\includegraphics[height=0.302\linewidth]{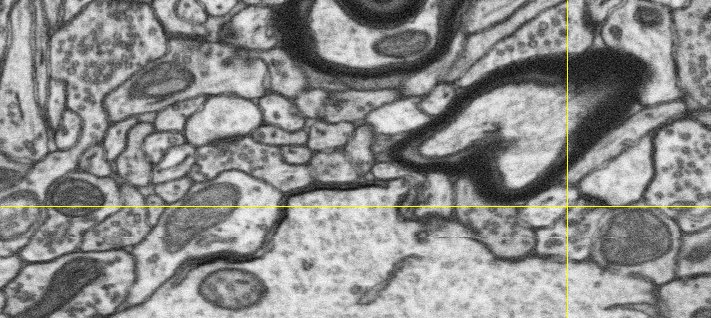}\\[-0.1cm]
  volume cut&(xz)\\
\end{tabular}
 \end{center}
\caption{Interface  of  the  FIJI   Visualization  API~\cite{Schmid10},  which  is
  extensively used to interact with 3D  image stacks. The user is presented with
  three  orthogonal planar  slices of  the stack.  While effective  when working
  slice  by slice,  this is  extremely cumbersome  for random  access to  voxels
  anywhere in the 3D stack, which is what a naive AL implementation
  would require.  
  }
\label{fig:01-fijiinterface}
\end{figure}

More   specifically,    3D   stacks   such   as    those   depicted   by
  Fig.~\ref{fig:01-fijiinterface}  are common  in the  biomedical field  and are
  particularly challenging, in part because it is difficult both to develop effective interfaces to visualize the huge image data and
  for users to quickly  figure out what they are looking at.   In this paper, we
  will therefore focus  on image volumes  but  the  techniques we will
  discuss are nevertheless also applicable to regular 2D images by treating them as stacks of
  height one.

\begin{figure}[h]
  \begin{center}
    \begin{tabular}{cc}
      \includegraphics[width=0.5\linewidth]{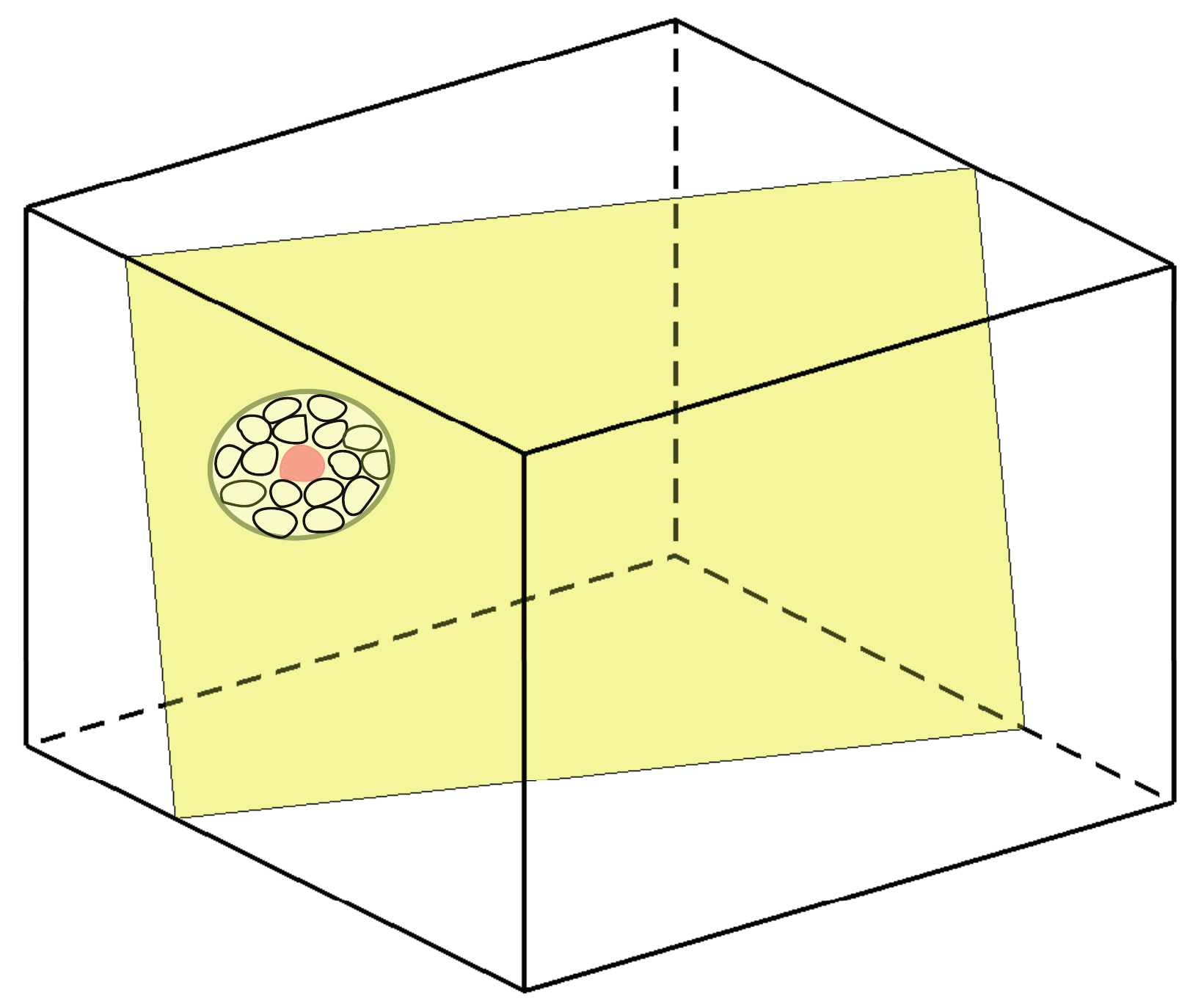}&
      \includegraphics[width=0.4\linewidth]{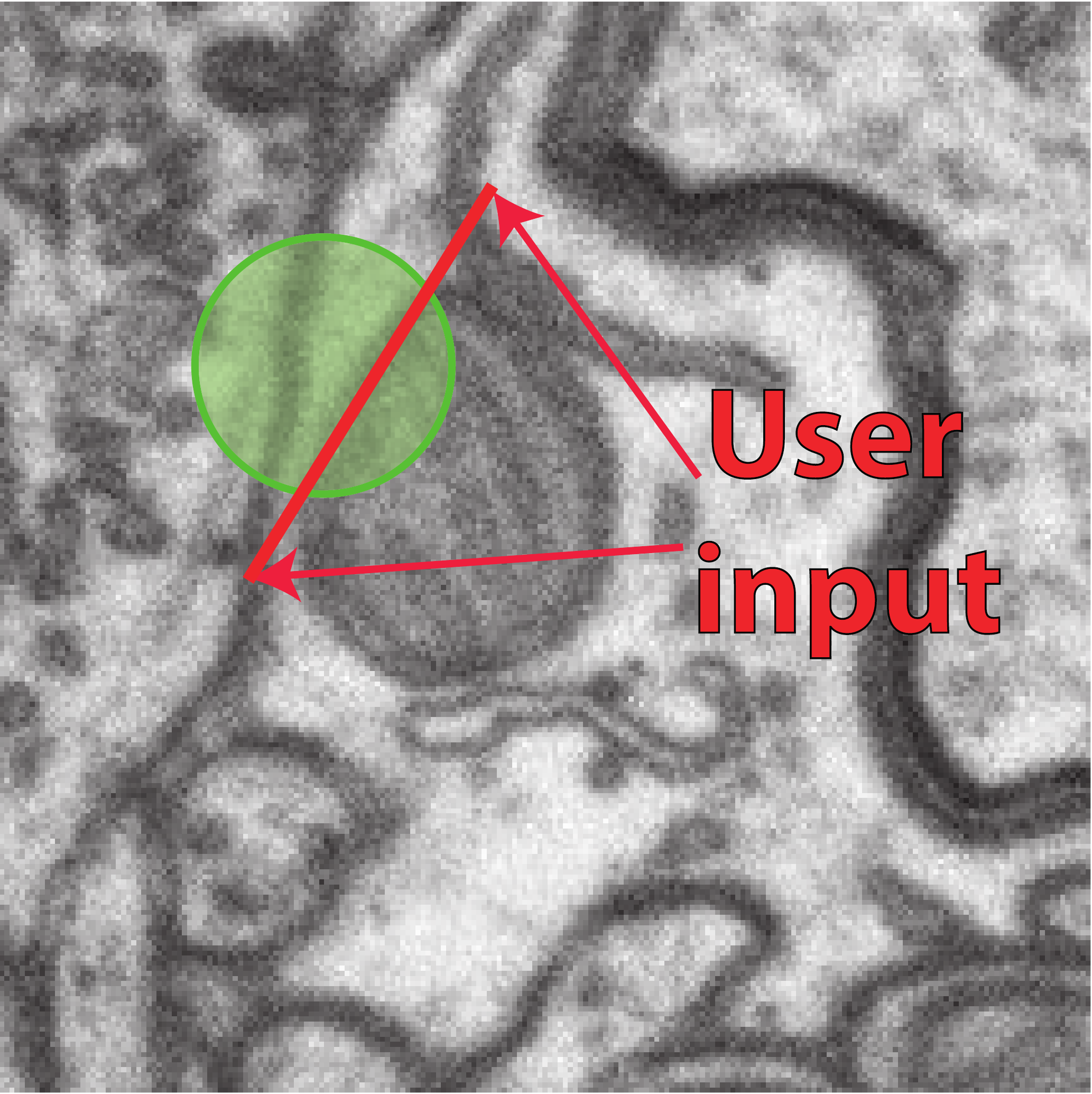}\\
      (a)&(b)
    \end{tabular}
  \end{center}
  \caption{Our approach to  annotation.  The system selects an  optimal plane in
    an arbitrary orientation---as opposed to  only xy, xz, and yz---and presents
    the user  with a  patch that is  easy to annotate.   (a) The  annotated area
    shown as  part of the full  3D stack.  (b)  The planar patch the  user would
    see.  It  could be annotated  by clicking twice  to specify the  red segment
    that forms  the boundary between the  inside and outside of  a target object
    within the green circle. Best viewed in color.}
\label{fig:01-planeInterface}
\end{figure}

With this,  we introduce here a novel approach  to AL that
is geared towards  segmenting 3D image volumes and also applicable to ordinary  2D  images.   By design, it takes  into  account  geometric
constraints to  which regions  should obey  and  makes the  annotation process convenient. Our contribution hence is twofold:
\begin{itemize}

  \item  We introduce  a way  to exploit  geometric priors  to more  effectively
    select the image data the expert user is asked to annotate.

  \item We  streamline the annotation  process in  3D volumes so  that annotating
    them is no  more cumbersome than annotating ordinary 2D  images, as depicted
    by  Fig.~\ref{fig:01-planeInterface}.

\end{itemize}

In the  remainder of this  paper, we first  review current approaches  to AL  and discuss  why  they are  not necessarily  the  most effective  when
dealing with pixels  and voxels. We then  give a short overview  of our approach
and  discuss in  more  details how  we  use geometric  priors  and simplify  the
annotation process.   Finally, we compare  our results against those  of 
accepted baselines and state-of-the-art techniques.

\section{Related Work and Motivation}
\label{sec:related}

In  this paper,  we  are  concerned with  situations  where  domain experts  are
available to annotate images.  However, their  time is limited and expensive. We
would  therefore like  to exploit  it  as effectively  as possible.   In such  a
scenario, AL~\cite{Settles10} is a technique of  choice as it tries to determine the smallest possible set of  training samples to annotate for effective model instantiation.

In practice, almost any classification scheme can be incorporated into an AL   framework.    For   image    processing   purposes,   that   includes
SVMs~\cite{Joshi09},  Conditional  Random Fields  \cite{Vezhnevets12},  Gaussian
Processes \cite{Kapoor07} and Random Forests \cite{Maiora12}. Typical strategies
for query selection rely    on     uncertainty     sampling~\cite{Joshi09},
query-by-committee~\cite{GiladBachrach05,Iglesias11},       expected       model
change~\cite{Settles08,Sznitman10,Vezhnevets12},  or measuring  information  in the  Fisher
matrix~\cite{Hoi06}.

These  techniques   have  been   used  for  tasks   such  as   Natural  Language
Processing~\cite{Lewis94,Tong02,Olsson09},                                 Image
Classification~\cite{Joshi09,Hoi06},                 and                Semantic
Segmentation~\cite{Vezhnevets12,Iglesias11}.  However,  selection strategies are
rarely  designed  to  take  advantage   of  image  specificities  when  labeling
individual pixels or voxels, such as the fact that neighboring ones tend to have
the same  labels or  that boundaries  between similarly  labeled ones  are often
smooth.  The  segmentation methods presented in~\cite{Li11,Iglesias11}  do however take
such geometric constraints into account at classifier level but not in AL query selection, as
we do.

Similarly,                            batch-mode
selection~\cite{Settles11,Hoi06,Settles08,Elhamifar13}    has   become    a
standard way to increase  the efficiency by  asking the expert to
annotate  more than  one  sample at  a time~\cite{Olabarriaga01,Altaie14}.   But
again, this  has been mostly investigated  in terms of semantic  queries without
due consideration to the fact that, in  images, it is much easier for annotators
to quickly label many samples in a localized image patch than having to annotate
random          image          locations.          In          3D          image
volumes~\cite{Li11,Iglesias11,Gordillo13},  it  is  even  more  important  to
provide the  annotator with  a patch in  a well-defined plane,  such as  the one
shown in Fig.~\ref{fig:01-planeInterface}, rather  than having him move randomly
in  a complicated  3D volume,  which is  extremely cumbersome  using current  3D image
display   tools    such   as   the    popular   FIJI   platform    depicted   by
Fig.~\ref{fig:01-fijiinterface}.  The technique of~\cite{Top11b} is an exception
in  that it  asks users  to label  objects  of interest  in a  plane of  maximum
uncertainty.   Our approach is similar, but  it  also  incorporates
geometric constraints in query selection and  as we show in the result section,
it outperforms the earlier method.


\section{Approach}

We  begin by  broadly outlining  our framework,  which is  set in  a traditional
AL context.  That is,  we  wish to  train  a classifier  for
segmentation purposes,  but have initially  only few labeled and  many unlabeled
training samples at our disposal.

Since segmentation of 3D volumes  is computationally expensive, supervoxels have
been  extensively used  to speed  up the  process~\cite{Andres08,Lucchi11b}.  We
therefore formulate  our problem in  terms of classifying supervoxels  as either
part of a target  object or not.  As such, we start  by oversegmenting the image
using  the  SLIC algorithm~\cite{Achanta12}  and  computing  for each  resulting
supervoxel  $s_i$  a feature  vector  $\bx_i$.   Note that  SLIC
  superpixels/supervoxels are always roughly circular/spherical, which allows us
  to characterize them by their center and radius.

Our AL  problem thus involves  iteratively finding  the next set  of supervoxels
that  should be  labeled by  an expert  to improve  segmentation performance  as
quickly as possible.  To this end, our algorithm proceeds as follows:
\begin{enumerate}
  
\item Using  the already manually labeled  voxels $S_L$, we  train a task specific  classifier and use it  to predict for
  all remaining voxels $S_U$ the probability of being foreground or background.
  
\item Next,  we score unlabeled  supervoxels on the  basis of a  novel uncertainty
  function   that  combines   traditional  Feature   Uncertainty  with   Geometric
  Uncertainty.   Estimating the  former  usually involves  feeding the  features
  attributed to each  supervoxel to the previously trained  classifier.   To
  compute  the latter,  we look  at the  uncertainty of  the label  that can  be
  inferred  based  on a  supervoxel's  distance  to  its neighbors  and  their
  predicted labels.  By doing so, we effectively capture the constraints imposed
  by the local smoothness of the image data.
  
\item We then automatically select the best  plane through the 3D image volume in
  which     to     label     additional      samples,     as     depicted     in
  Fig.~\ref{fig:01-planeInterface}.  The expert can then effortlessly label the
  supervoxels from a circle in the  selected plane by defining a line separating
  target from non-target regions. This removes  the need to examine the relevant
  image     data    from     multiple    perspectives,     as    depicted     in
  Fig.~\ref{fig:01-fijiinterface}, and simplifies the labeling task.

\end{enumerate}
The process is then repeated.  In Sec.~\ref{sec:GeomActive}, we will discuss the
second    step    and    in     Sec.~\ref{sec:BatchGeom}    the    third.     In
Sec.~\ref{sec:exp}, we will demonstrate  that this pipeline yields faster
learning rates than competing approaches.


\section{Geometry-Based Active Learning}
\label{sec:GeomActive}

Most AL  methods were  developed for  general tasks  and operate  exclusively in
feature  space, thus  ignoring the  geometric properties  of images  and more
specifically their  geometric consistency.  To  remedy this, we introduce
the concept of Geometric  Uncertainty and then show how to combine it with more
traditional Feature Uncertainty.

Our basic insight is that supervoxels that  are assigned a label other than that
of their  neighbors ought to  be considered more  carefully than those  that are
assigned the  same labels. In other  words, under the assumption  that neighbors
more often  than not have  identical labels, the  chance of the  assigment being
wrong is higher. This is what we  refer to as {\it Geometric Uncertainty} and we
now  formalize it. 

\subsection{Feature Uncertainty}
\label{sec:FeatUncert}

For each supervoxel $s_i$ and  each class $\hat{y}$, let $p_{\theta}(y_i=\hat{y}
| \bx_i)$ be the probability that its class $y_i$ is  $\hat{y}$, given the
corresponding feature vector $\bx_i$. In this work, we will assume that this probability can be computed by means of a classifier which has been trained using parameters $\theta$ and we take $\hat{y}$ to be 1 if the supervoxel belongs to the foreground and 0 otherwise \footnote{Extensions to multiclass cases can be similarly derived.}. In  many  AL  algorithms,  the uncertainty of this prediction is taken to be the Shannon entropy
\begin{equation}
 H^{\theta}_i = -\sum_{\hat{y} \in \{0,1\}} p_{\theta}(y_i=\hat{y} | \bx_i) \log
 p_{\theta}(y_i=\hat{y} | \bx_i). \; \;
 \label{eq:FeatEntropy}
\end{equation}
We  will refer to this uncertainty estimate as the {\it Feature Uncertainty}.

\subsection{Geometric Uncertainty}
\label{sec:GeomUncert}

Note that estimating  the uncertainty as described above  completely ignores the
correlations  between neighboring  supervoxels. To account  for  them, we  can
estimate the  entropy of a  different probability, specifically  the probability
that  supervoxel $s_i$ belongs  to class $\hat{y}$  given the
classifier predictions of its neighbors and which we denote $p_G(y_i=\hat{y})$.

\begin{figure}[t]
\begin{center}
   \includegraphics[width=0.9\linewidth]{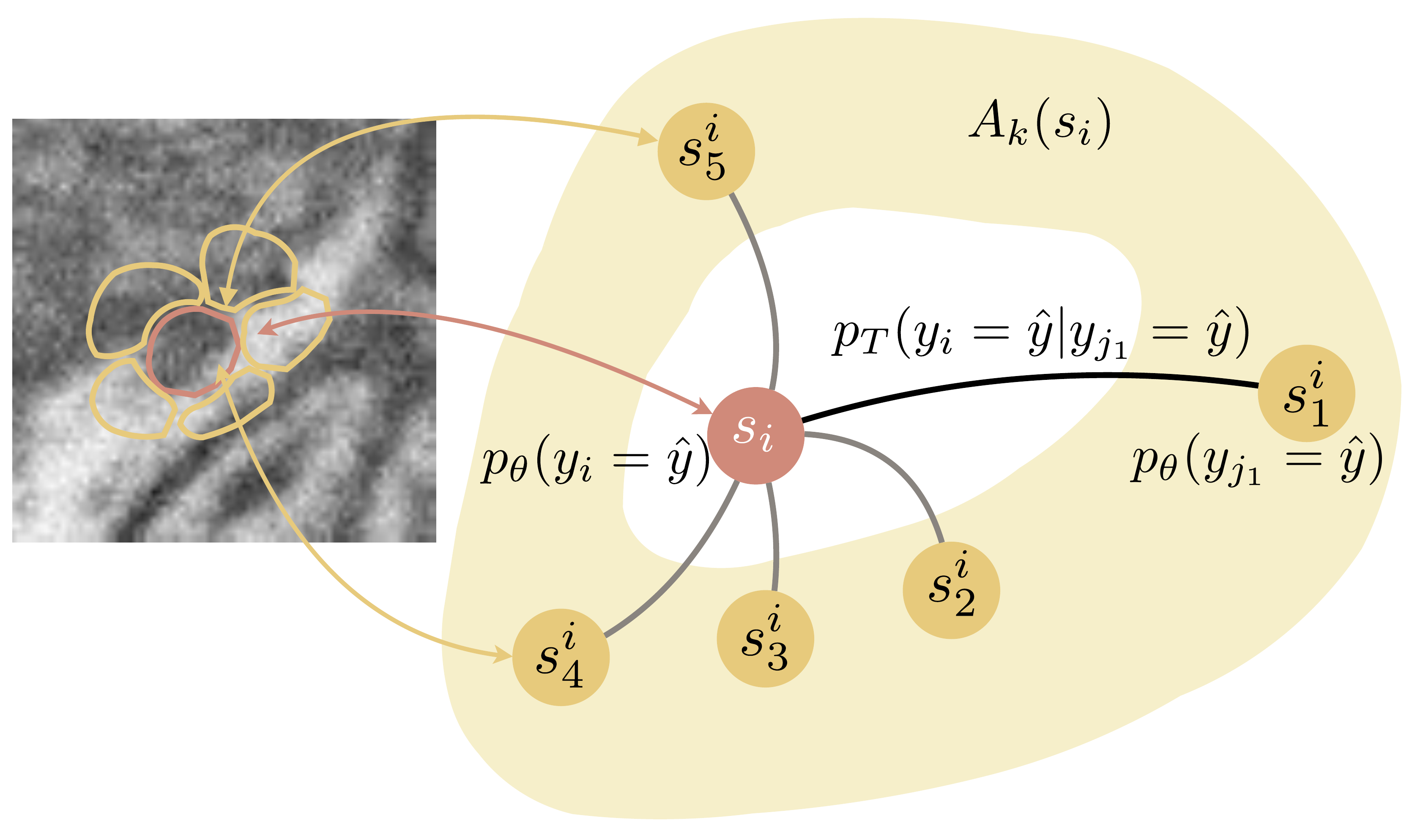}
\end{center}
   \caption{Image represented as a graph: we treat supervoxels as nodes in the graphs and edge weights between them reflect the probability of transition of the same label to a neighbour. 
   Supervoxel $s_i$ has $k$ neighbours from $A_k(i) = \{s_{1}^i, s_{2}^i, .., s_{k}^i\}$, 
$p_T(y_i=\hat{y}|y_j=\hat{y})$ is the probability of  node $s_i$ having  the same label as  node $s_j^i$, $p_{\theta}(y_i=\hat{y}|\bx_i)$ is the probability that $y_i$ , class of $s_i$, is $\hat{y}$, given only the corresponding feature vector $\bx_i$}
\label{fig:04-graph}
\end{figure}

To this  end, we treat  the supervoxels of  a single image  volume as nodes  of a
weighted  graph  $G$  whose  edges  connect neighboring  ones,  as  depicted  in
Fig.~\ref{fig:04-graph}. We let $A_k(s_i)=\{s^i_{1},s^i_{2}, .., s^i_{k}\}$ be the set of $k$ nearest neighbors of $s_i$ and assign a weight inversely proportional to the Euclidean distance between the voxel centers  to each one of the edges. For each node $s_i$, we normalize the weights of all incoming edges so that   their    sum   is   one    and   treat   this   as    the   probability $p_T(y_i=\hat{y}|y_j=\hat{y})$ of node $s_i$ having  the same label as node $s_j^i \in A_k(s_i)$. In  other words, the closer  two nodes are, the  more likely they are to have the same label. 



To infer $p_G(y_i=\hat{y})$, we then use the well-studied Random Walk strategy $G$~\cite{Lovasz93}, as it reflects  well our
smoothness  assumption and has been extensively used for image segmentation purposes~\cite{Grady06,Top11b}. Given  the  $p_T(y_i=\hat{y}|y_j=\hat{y})$  transition probabilities,  we  can compute the probabilities $p_G$ iteratively by initially taking $p^0_{G}(y_i=\hat{y}) $ to be
\begin{footnotesize}
  \begin{equation}
    p^0_{G}(y_i=\hat{y})                         =                        \sum_{s_j\in
  A_k(s_i)}p_T(y_i=\hat{y}|y_j=\hat{y})p_\theta(y_j=\hat{y}|\bx_j) \; \; ,
  \end{equation}
\end{footnotesize}
and then iteratively computing
\begin{footnotesize}
  \begin{equation}
  p^{\tau+1}_{G}(y_i=\hat{y})                     =                     \sum_{s_j\in
    A_k(s_i)}p_T(y_i=\hat{y}|y_j=\hat{y})p^{\tau}_{G}(y_j=\hat{y}) \; \; .
\label{eq:GeomProb}
  \end{equation}
\end{footnotesize}
\noindent

The  procedure describes  the  propagation  of labels  to  supervoxels from  its
neighborhood.  The number  of iterations $\tau_{max}$ defines the  radius of the
neighborhood involved  in the  computation of  $p_G$ for  $s_i$ and  encodes the
smoothness priors.

Given these probabilities, we can now take the {\it Geometric Uncertainty} to be
\begin{footnotesize}
\begin{equation}
 H^{G}_i = -\sum_{\hat{y} \in \{0,1\}} p_{G}(y_{i}=\hat{y}) \log
   p_{G}(y_i=\hat{y}) \; \; ,
\label{eq:GeomEntropy}
\end{equation}
\end{footnotesize}
as we did in Sec.~\ref{sec:FeatUncert} to estimate the Feature Uncertainty. 

\subsection{Combining Feature and Geometric Entropy}
\label{sec:CombinedUncert}

As discussed  above, from a trained classifier  we can thus
estimate  the  Feature  and  Geometric  Uncertainties. To use them  jointly, we should in   theory   estimate   the    joint   probability   distribution   $p_{\theta,G}(y_i=\hat{y}|x_i)$  and  the  corresponding   joint  entropy.   As  this  is computationally intractable in our model, we take advantage of the fact that the joint entropy is  upper bounded by the sum of  individual entropies $H^{\theta}$ and $H^{G}$. Thus,  for each supervoxel, we take the  {\it Combined Uncertainty}
to be
\begin{equation}
  H^{\theta,G}_i = H^{\theta}_i + H^G_i
  \label{eq:CombEntropy}
\end{equation}
\comment{\begin{multline}
s^* = \underset{s_i \in S_U}{\arg\max} \sum_{\hat{y}} p_{\theta}(y_i=\hat{y}|\bx_i) \log p_{\theta}(y_i=\hat{y}|\bx_i) +
\\
\sum_{\hat{y}} p_G(y_i=\hat{y}|\bx_i) \log p_G(y_i=\hat{y}|\bx_i) \; ,

\end{multline}}
that is, the upper bound of the joint entropy. 

In practice, using this measure means that supervoxels that individually receive uncertain predictions and are in areas of transition between foreground and background will be considered first.


\section{Batch-Mode Geometry Query Selection}
\label{sec:BatchGeom}

The   simplest  way   to  exploit   the  Combined   Uncertainty  introduced   in
Sec.~\ref{sec:CombinedUncert} would  be to  pick the most  uncertain supervoxel,
ask the expert to label it, retrain the classifier, and iterate. A more
effective way however is to find  appropriately-sized batches of  uncertain supervoxels
and  ask the  expert to  label them  all before  retraining the  classifier.  As
discussed  in   Sec.~\ref{sec:related},  this  is  referred   to  as  batch-mode
selection.  A naive implementation of this would force the user to randomly view
and annotate supervoxels  in the volume regardless of where they are, which would be  extremely cumbersome.

In this  section, we therefore  introduce an  approach to using  the uncertainty
measure to first select  a planar patch in 3D volumes and then  to allow the user
to quickly label positives and negatives within it, a shown in 
Fig.~\ref{fig:01-planeInterface}.

\begin{figure}[t]
\begin{center}
   \includegraphics[width=1\linewidth]{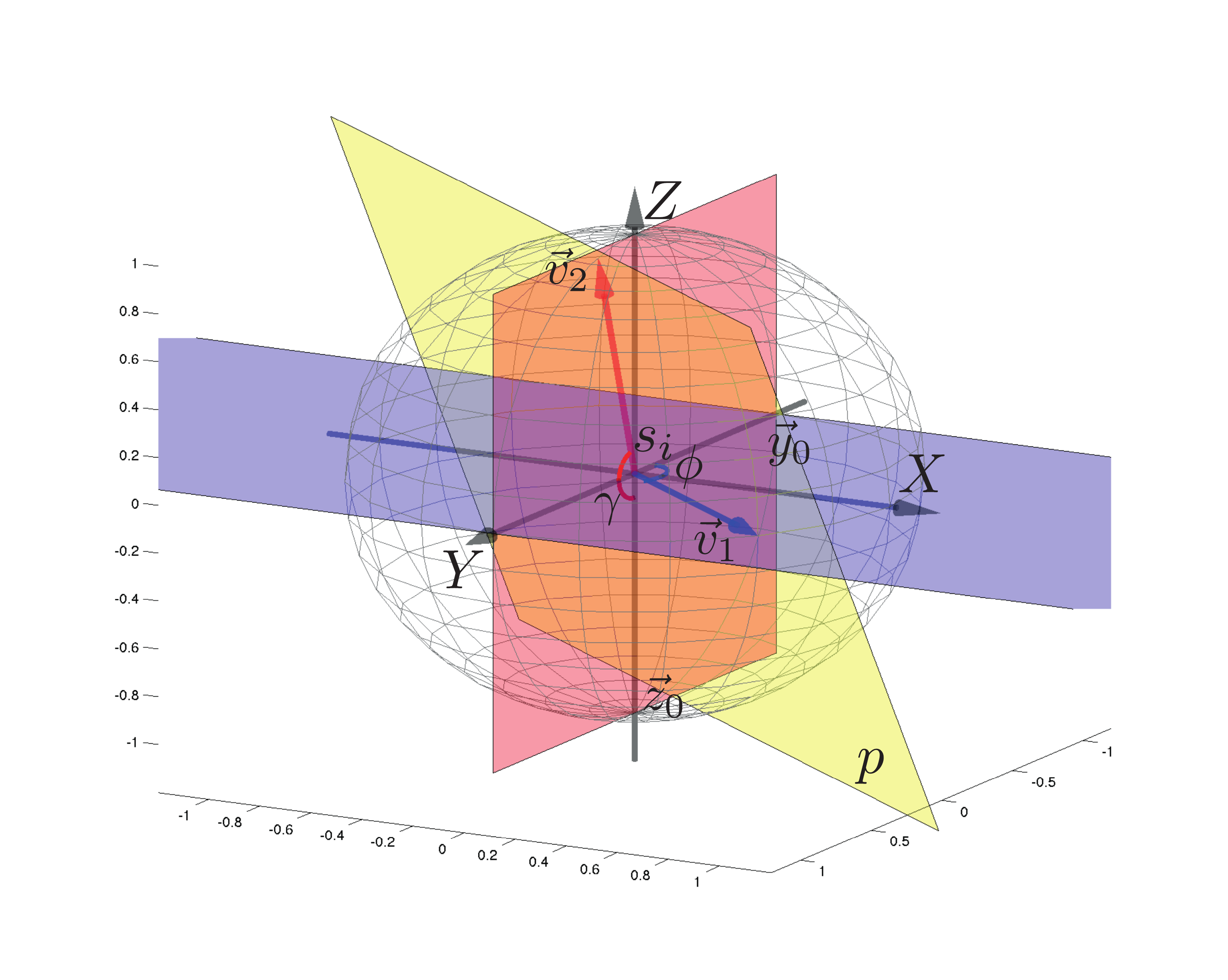}
\end{center}
   \caption{Coordinate system for defining planes. Plane $p_i$ (yellow) is defined by two angles $\phi$ -- intersection between plane $p$ and plane $X s_i Y$ (blue), $\gamma$ -- intersection between plane $p_i$ and plane $Y s_i Z$ (red). Best seen in color.}
   \label{fig:05-coordSys}
\end{figure}

Since we are working in 3D and there is no preferential orientation in the
data  to work  with, it  makes sense  to look  for spherical  regions where  the
uncertainty  is  maximal. However,  for  practical  reasons,  we only  want  the
annotator to  consider circular regions  within planar  patches such as  the one
depicted in  Fig.~\ref{fig:01-planeInterface} and Fig.~\ref{fig:05-coordSys}.  These  can be understood  as the
intersection of the sphere with a plane of arbitrary orientation.

Formally, let  us consider  a supervoxel  $s_i$ and let  $\mathcal{P}_i$ to be the set of planes bisecting  the image  volume and  containing the  center of  $s_i$.  Each
plane  $p_i \in  \mathcal{P}_i$  can  be parameterized  by  two  angles $\phi  \in
(0,\pi)$ and  $\gamma \in  (0,\pi)$ and  its origin  can be  taken to  be the
center of  $s_i$, as depicted by  Fig.~\ref{fig:05-coordSys}.  
In  addition, let $C^r_i(p_i)$ be the set  of supervoxels lying on
$p_i$ within distance $r\geq0$  to its origin,
shown in green in Fig.~\ref{fig:01-planeInterface}.

Recall   from  Sec.~\ref{sec:GeomActive},  that  we   have  defined  the
uncertainty   of   a  supervoxel   as   either   the  Feature   Uncertainty   of
Eq.~\eqref{eq:FeatEntropy},       the       Geometric       Uncertainty       of
Eq.~\eqref{eq:GeomEntropy},     or      the     Combined      Uncertainty     of
Eq.~\eqref{eq:CombEntropy}. In  other words, we  can associate each  $s_i$ with an
uncertainty value $u_i  \geq 0$ in one  of three ways. Whichever  way we choose,
finding the circle of maximal uncertainty can be formulated as finding
\begin{equation}
p^\ast_i =  (\phi^\ast,\gamma^\ast) =  \argmax_{p_i \in  \mathcal{P}_i} \sum_{s_j\in
  C^r_i(p_i)} u_j \;\; .
\label{eq:optimization}
\end{equation}
\noindent
In practice, we find the  most uncertain plane $p^\ast_i$  for the $t$ most uncertain supervoxels $s_i$ and
present the overall uncertain plane $p^*$ to the annotator. Since $u_i\geq0$ and Eq.~\eqref{eq:optimization} is linear in $u_i$, we designed a branch-and-bound approach to solving Eq.~\ref{eq:optimization}. It uses a bounding function to quickly eliminate entire parts  of the parameter space until it is reduced to  a singleton.  By  contrast to an  exhaustive search that  would be excruciatingly slow,  our current  MATLAB implementation  on the 10 images of resolution  $176\times 170\times 220$ of MRI dataset of Sec.~\ref{sec:MRI} takes $0.12$s per plane selection. This means that a C implementation would be real-time, which is critical to  such an  interactive method  to being accepted  by users. We discuss our implementation in more details in the supplementary material.

Note that when the radius $r=0$, this reduces to what single-supervoxel labeling does.  By contrast,  for  $r  > 0$,  this  allows annotation  of many  uncertain supervoxels  with a few  mouse clicks,  as will  be discussed further in Sec.~\ref{sec:exp}.
Although planar selection can be applied to any type of uncertainty value, we believe that it is the most beneficial when combined with Geometric Uncertainty as the latter already takes into account the most uncertain regions instead of isolated supervoxels.

\section{Experiments}
\label{sec:exp}

\begin{figure*}[h]
\begin{center}
\begin{tabular}{ccc}
\includegraphics[width=0.32\linewidth]{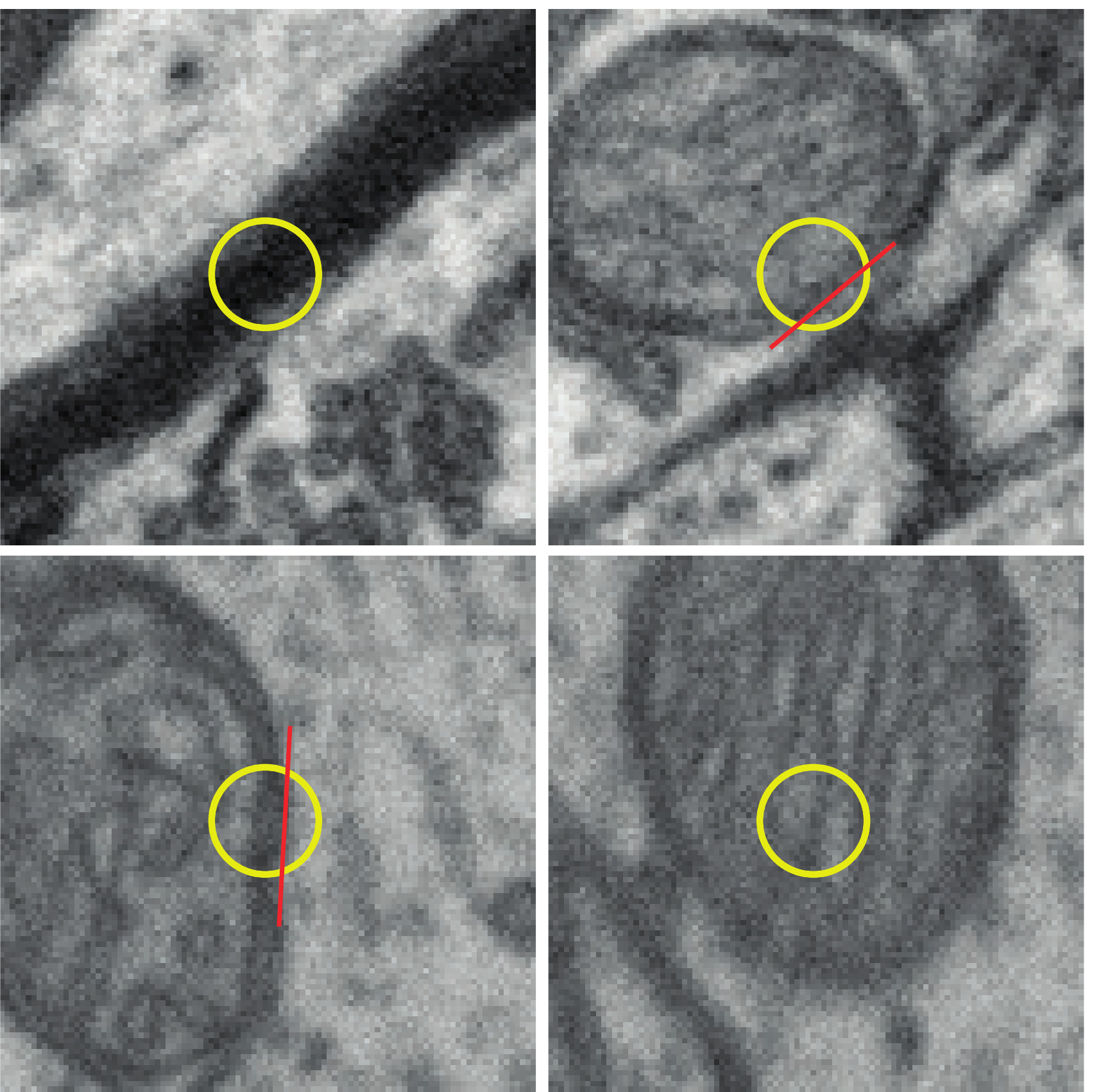} &
\includegraphics[width=0.32\linewidth]{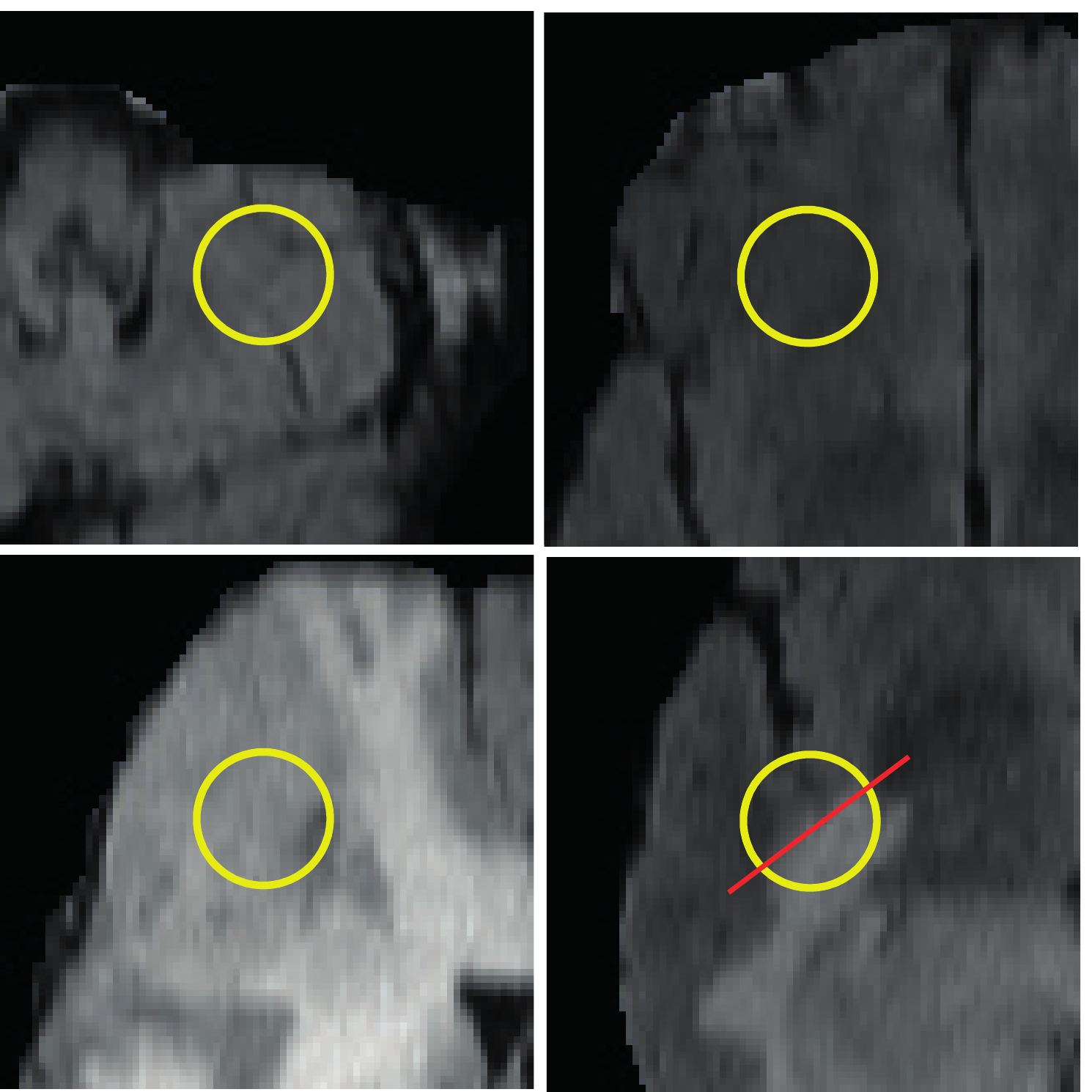} &
\includegraphics[width=0.32\linewidth]{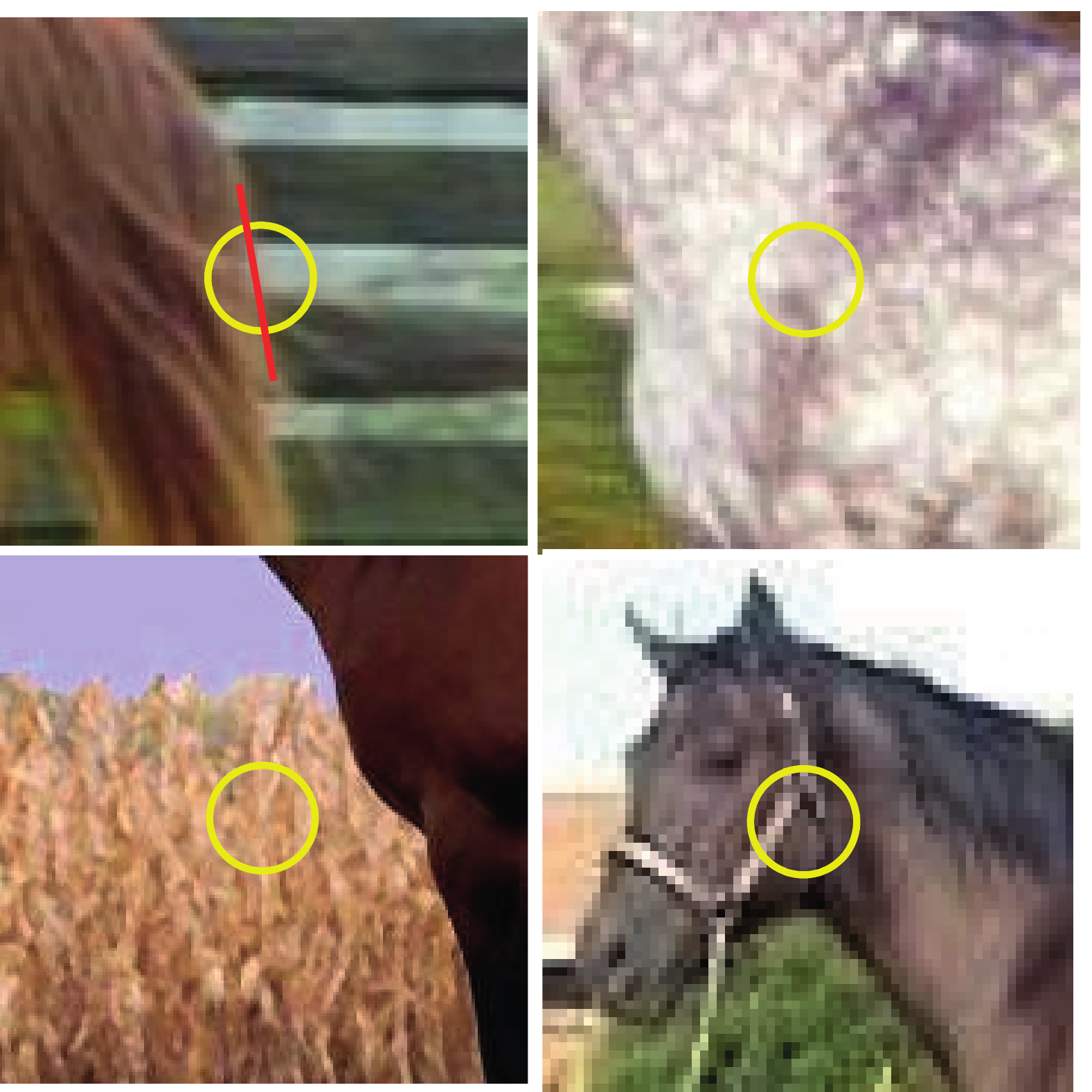} \\
(a)&(b)&(c)
\end{tabular}
\end{center}
   \caption{Circular patches  to be annotated  by the expert highlighted  by the
     yellow  circle in (a)  Electron Microscopy  data, (b)  MRI data,  and (c)  natural
     images.     The   patches    can   be    entirely   foreground,    entirely
     background. Alternatively,  the boundary  between foreground  an background
     within the  patch can be  indicated by tracing a  red line segment.  In all
     cases, that would require at most two mouse clicks.}
\label{fig:06-selection}
\end{figure*}

In this  section, we evaluate our  full approach both on  two different Electron
Microscopy (EM) datasets and a Magnetic Resonance Imaging (MRI) one.  
We then  demonstrate  that a  simplified  version  is  effective for  natural  2D Images.

\subsection{Setup and Parameters}
\label{sec:setup}

For  all   our  experiments,  we   used  Boosted  Trees  selected   by  Gradient
Boosting~\cite{Sznitman13a,Becker13b} as our  underlying classifier.  Given that
during early  AL iterations rounds,  only limited  amounts of training  data are
available,  we  limit  the depth  of  our  trees  to  2 to  avoid  over-fitting.
Following standard practice, individual trees are optimized using $40\%-60\%$ of
the  available training  data chosen  at random  and $10$  to $40$  features are
explored  per   split.   We  set  the   number  $k$  of  nearest   neighbors  of
Section~\ref{sec:GeomUncert}   to  be   the  number   of  immediately   adjacent
supervoxels  on average,  which is  between 7  and 15  depending on  the
  resolution of the image and  size of supervoxels.  
  However, experiments showed  that the algorithm  is not very  sensitive to the  choice of
this parameter.  We restrict the size of each planar patch to be small enough to
contain typically not  more than part of  one object of interest.   To this end,
the we take the radius $r$  of Section~\ref{sec:BatchGeom} to be between 10 and 15, which     yields    patches     such     as     those    depicted     by
  Fig.~\ref{fig:06-selection}.

\vspace{-0.2cm}
\paragraph{Baselines.}

For  each dataset,  we  compare  our approach  against  several baselines.   The
simplest is Random  Sampling (\RS{}), that is, randomly selecting  samples to be
labeled.  It  serves to  gauge the  difficulty of  the segmentation  problem and
quantify the improvement brought by the more elaborate strategies.

The  next simplest,  but  widely  accepted approach  is  to perform  Uncertainty
Sampling~\cite{Elhamifar13,Long13}  of  supervoxel   by  using  the  uncertainty
measures of Section~\ref{sec:GeomActive}.  Let  $H_i^U$ be the uncertainty score
we use in a specific experiment.  The strategy then is to select
\begin{equation}
 s^*= \underset{s_i \in S_U}{\arg\max} (H_i^U). \; \;
\end{equation} 
We  will  refer  to  this  as  \FU{}  when  using  the  Feature  Uncertainty  of
Eq.~\ref{eq:FeatEntropy} and  as \CU{}  when using  the Combined  Uncertainty of
Eq.~\ref{eq:CombEntropy}.   For  the  Random   Walk,  iterative  procedure  with
$\tau_{max}=20$ leads to high learning  rates in our applications.  Finally, the
most sophisticated  approach is to  use Batch-Mode Geometry Query  Selection, as
described  in  Sec.~\ref{sec:BatchGeom},  in  conjunction  with  either  Feature
Uncertainty  or  Combined Uncertainty.   We  will  refer  to the  two  resulting
strategies as \PFU{} and \PCU{},  respectively.  Both plane selection strategies
are using $t=5$ best supervoxels in  the optimization.  Further increase of this
value didn't demonstrate significant growth of the learning rate.

Fig.~\ref{fig:01-planeInterface},~\ref{fig:06-selection} jointly  depict what a potential user
would see for \PFU{} and \PCU{} given a small enough patch radius. Given a well
designed interface,  it will typically  require to click  only once or  twice to
provide the required feedback (see Fig.~\ref{fig:06-selection}).  In our performance evaluation, we will therefore estimate that each  intervention of the user for $\PFU{}$  and $\PCU{}$ requires
two clicks whereas for $\RS{}$, $\FU{}$,  and $\CU{}$ it requires only one.  So,
for  the method  comparison  we  measure annotation  effort  as  1 for  $\RS{}$,
$\FU{}$, and $\CU{}$ and as 2 for $\PFU{}$ and $\PCU{}$.

Note that $\PFU{}$ is similar in spirit to the approach of~\cite{Top11b} and can
therefore be taken as a good indicator of how this other method would perform on
our data.   However, unlike~\cite{Top11b}, we do  not require user to  label the
whole plane and keep our suggested interface for a fairer comparison.

\begin{figure}[t]
\begin{center}
\begin{tabular}{c}
   \includegraphics[width=0.7\linewidth]{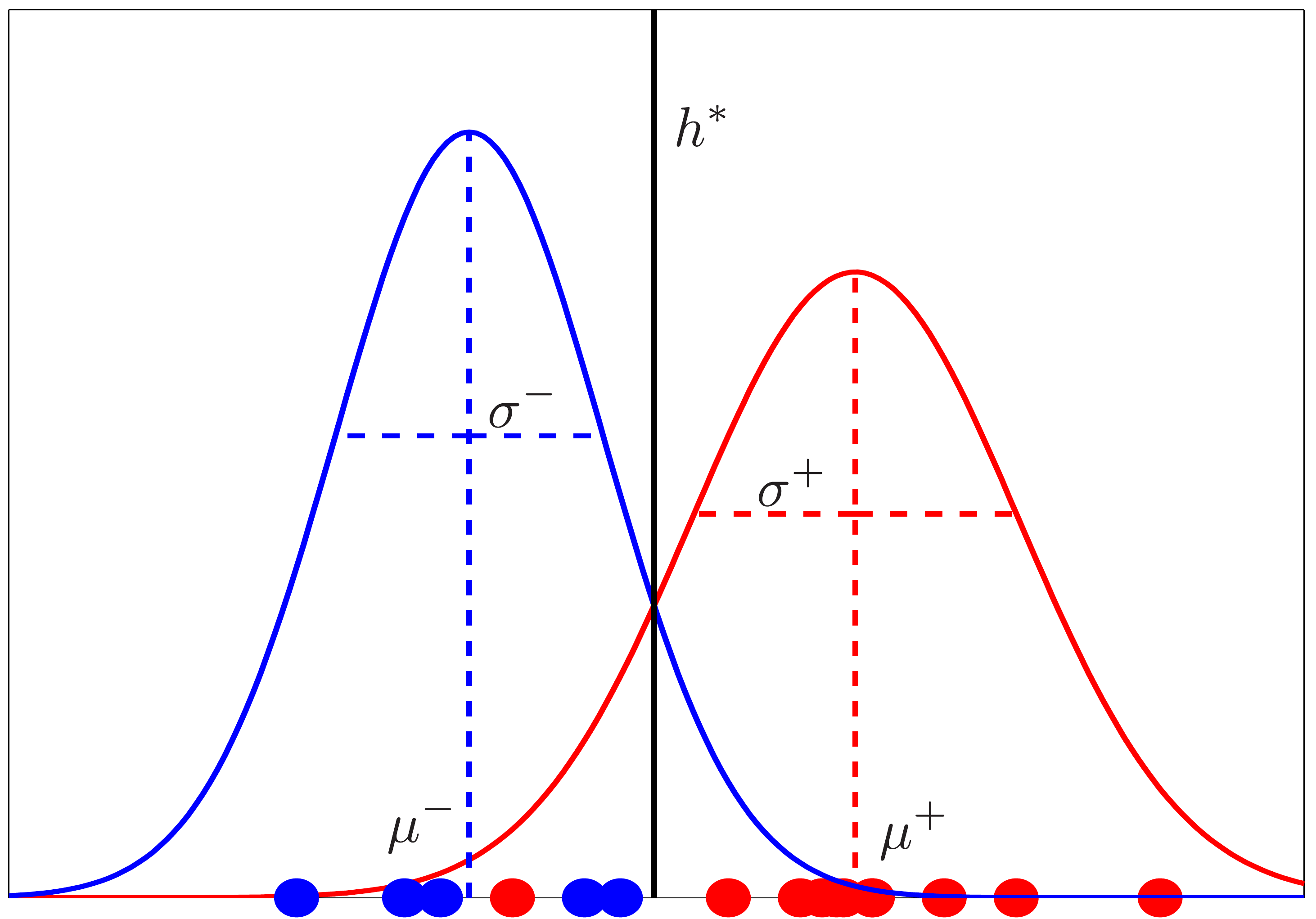}\\
  (a) \\
   \includegraphics[width=0.97\linewidth]{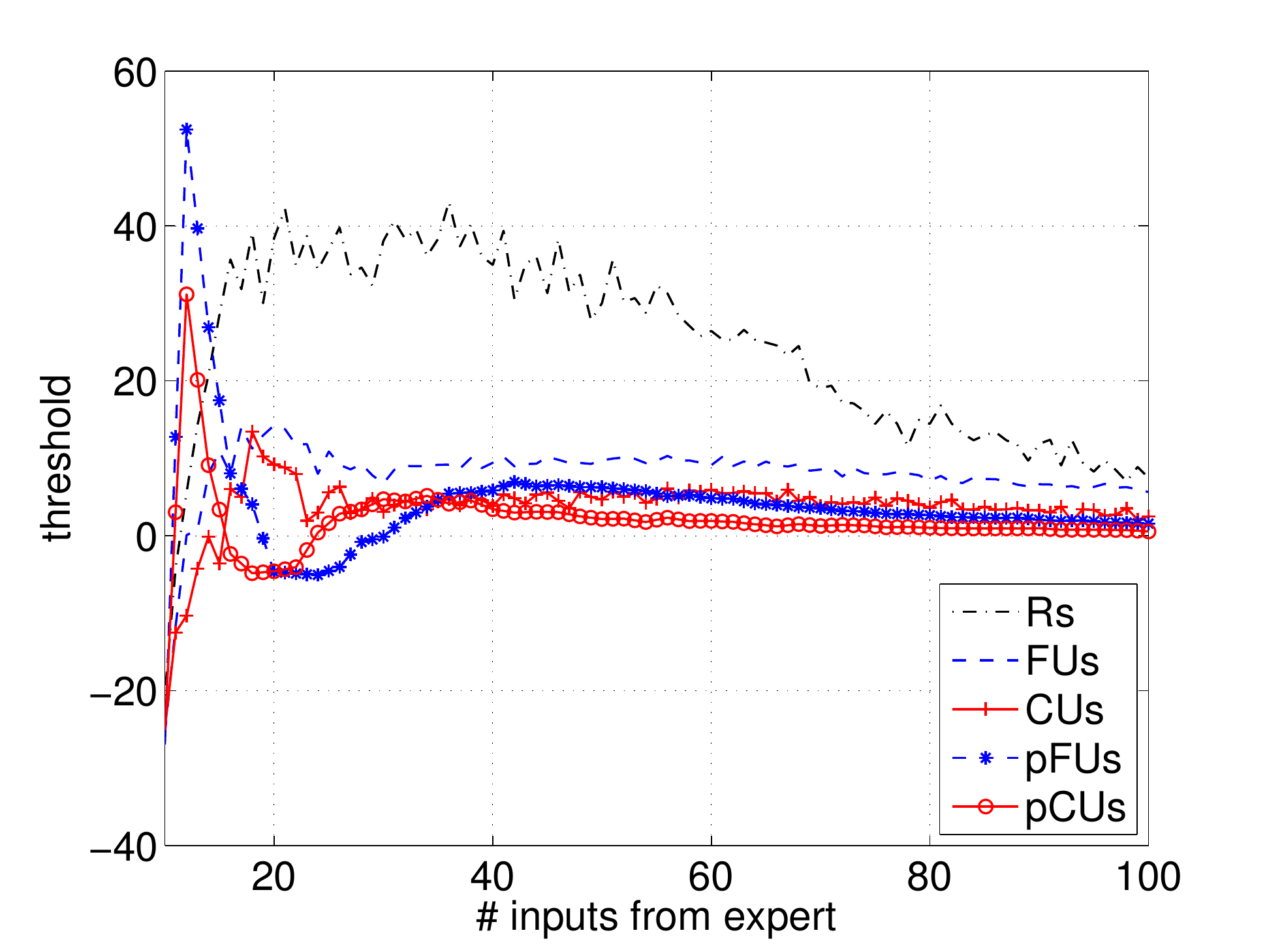}\\[-0.1cm]
  (b)\\
\end{tabular}
\end{center}
   \caption{(a) Estimate mean and standard deviation for classifier scores of positive class datapoints (red) -- $\mu^+$ and $\sigma^+$ and negative class datapoints (blue) -- $\mu^-$, $\sigma^-$, and fit 2 Gaussian distributions. Given their pdf estimate optimal Bayesian error with threshold $h^*$.
   (b) Adaptive Thresholding convergence rate of classifier threshold for different AL strategies.}
\label{fig:06-threshold}
\end{figure}

\vspace{-0.2cm}
\paragraph{Adpative Thresholding.}
Recall  from  Section~\ref{sec:FeatUncert}   that  for  all  the
  approaches  discussed here,  the  probability of a  supervoxel
  being foreground is  computed as $p_{\theta}(y_i=1 | x_i)=(1+\exp^{-2 \cdot (F-h)})^{-1}$, where $F$ is the output of the classifier and $h$ is the threshold ~\cite{Hastie01}.  Usually, the threshold  is chosen by
  cross-validation but this strategy may be  misleading or not even possible for
  AL.  We therefore assume that the scores of training samples in each class are
  Gaussian distributed with unknown parameters  $\mu$ and $\sigma$. We then find
  an optimal threshold $h^*$ by fitting  Gaussian distributions to the scores of
  positive and negative classes and choosing  the value that yields the smallest
  Bayesian error,  as depicted  by Fig.~\ref{fig:06-threshold}(a).  We  refer to
  this  approach as  {\it  Adaptive Thresholding}  and  we use  it  for all  our
  experiments. Fig.~\ref{fig:06-threshold}(b) depicts the  value of the selected
  threshold as increasing amounts of  annotated data become available. Note that
  different strategies yield varying convergence speeds and that the plane-based
  strategies ($\PFU{}$ and $\PCU{}$) converge fastest to a stable value.

\vspace{-0.2cm}  
\paragraph{Experimental Protocol.}

In all cases,  we start with 5  positive and 5 negative  labeled supervoxels and
perform AL iterations until we receive 100 inputs from the user.  Each  method starts with  the same random  subset of
samples  and  each  experiment  is  repeated $N=40$  times.   We  will
  therefore plot not only accuracy results but also indicate the variance of these results.

Fully annotated volumes of ground truth are  available for us and we use them to
simulate the expert's  intervention in our experiments.  We  detail the specific
features we used for EM, MRI, and natural images below.

\subsection{Results on EM data}

\begin{figure}
 \begin{center}
\begin{tabular}{c}
  \hspace{-0.3cm}\includegraphics[width=0.7\linewidth]{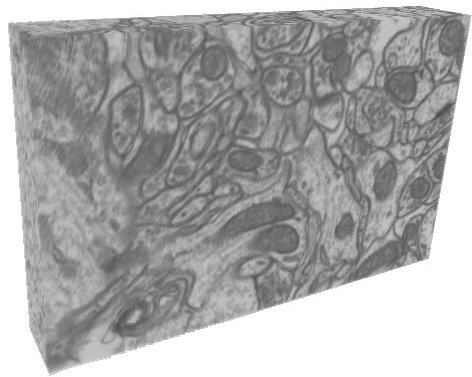} \\
  (a) \\
  \hspace{-0.3cm}\includegraphics[width=0.9\linewidth]{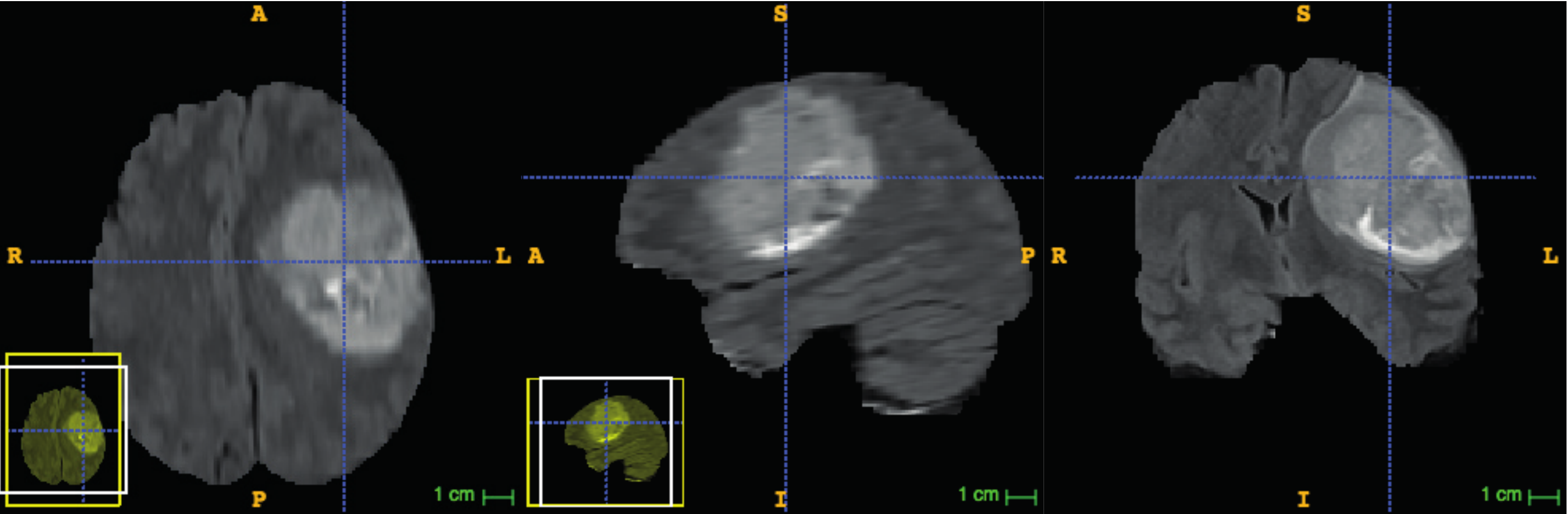}\\[-0.1cm]
  (b)\\
\end{tabular}
 \end{center}
\vspace{-0.3cm}
\caption{Examples of 3D datasets. a) Hippocampus volume for mitochondria segmentation b) MRI data for tumor segmentation (Flair image).}
\vspace{-0.3cm}
\label{fig:06-datasets}
\end{figure}

Here, we work with  two 3D Electron Microscopy stacks of  rat neural tissue, one
from the striatum  and the other from  the hippocampus.  One stack  of size $318
\times 711 \times 422$  ($165 \times 1024 \times 653$ for hippocampus) is  used for training and
another stack of size $318 \times 711 \times 450$ ($165 \times 1024 \times 883$)
is  used to  evaluate the  performance.  Their  resolution is  5nm in  all three
spatial orientations.  The slices  of Fig.~\ref{fig:01-fijiinterface} as well as
patches  in the  upper  row  in Fig.~\ref{fig:06-selection}  (a)  come from  the
striatum  and  hippocampus volume is shown in Fig.~\ref{fig:06-datasets} (a) with its patches  shown  in  the   lower  row  of
Fig.~\ref{fig:06-selection} (a).  

The task is to segment mitochondria, which are the intracellular structures that
supply the  cell with its energy  and are of great  interest to neuroscientists.
It is extremely laborious to annotate sufficient amounts of training data for
learning segmentation algorithms to work satisfactorily.  Furthermore, different
brain areas  have different characteristics, which  means that the task  must be
repeated often.   The features we feed  our Boosted Trees rely  on local texture
and  shape information  using ray  descriptors  and intensity  histograms as  in
~\cite{Lucchi11b}.
\comment{Answer comment from my readers "Is there any prove that including geometry in the classifier won't give the same result?". 
Note, that in these examples some geometrical context is already included in the features that are fed into classifier. 
However, the experiments prove that this indirect geometry in AL can be further improved by its explicit usage in query selection.
Recall that most works on AL in segmentation ~\cite{Li11,Iglesias11} only use geometrical constraints in the classifier.}

\begin{figure*}[h]
\begin{center}
\begin{tabular}{cc}
\includegraphics[width=0.4\linewidth]{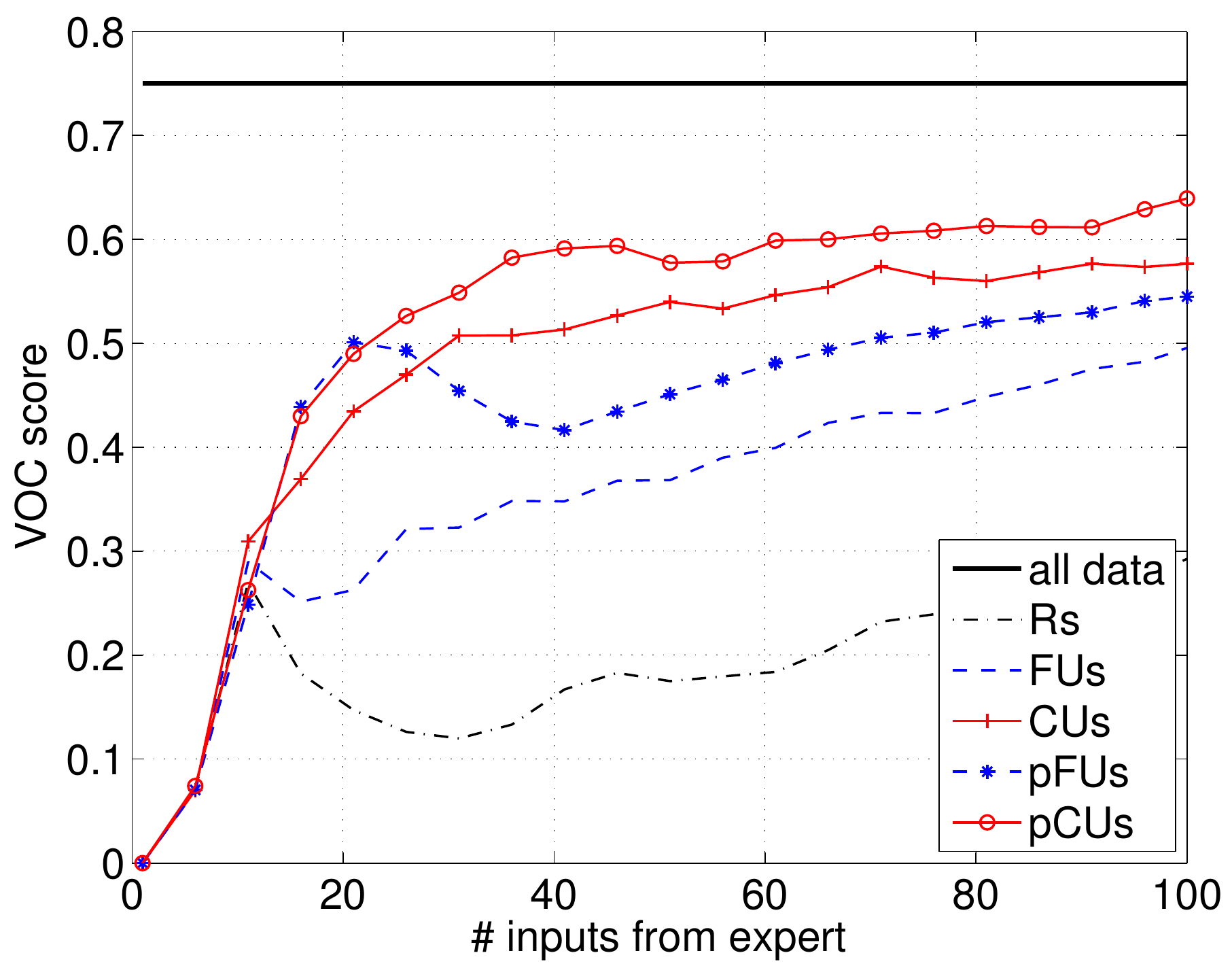} &
\includegraphics[width=0.4\linewidth]{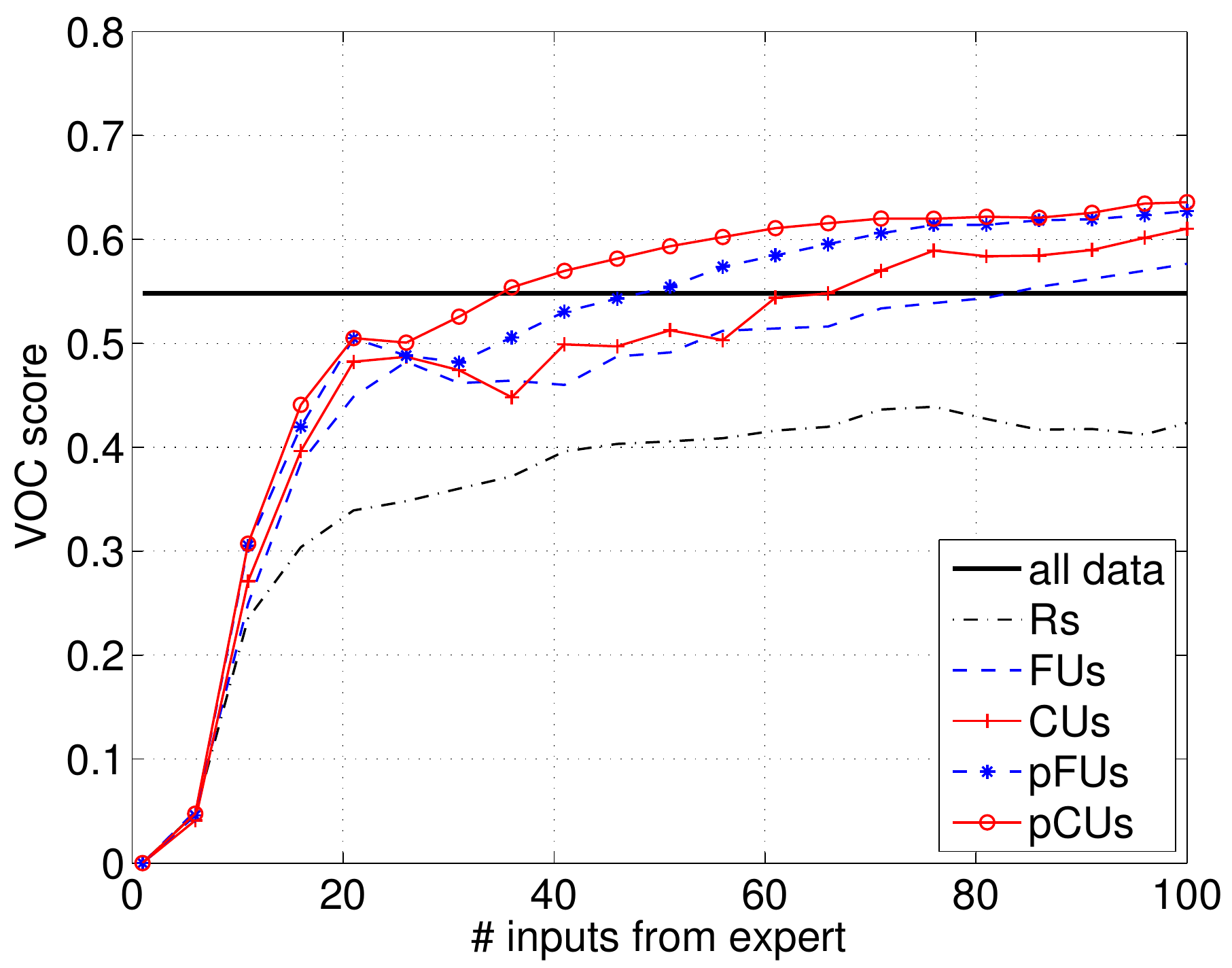}\\
(a)&(b)
\end{tabular}
\end{center}
   \caption{Comparison of various AL strategies for mitochondria segmentation. Left: striatum dataset, right: hippocampus dataset.}
\label{fig:emresults}
\end{figure*}
\begin{center}
\begin{table}[h]

  \begin{tabular}{|l||l|l|l|l| }
    \hline 
    Dataset & $\FU$ & $\CU$ & $\PFU$ & $\PCU$ \\ \hline \hline
    Hippocampus & 0.1172 & 0.1009 & 0.0848 & {\bf 0.0698} \\ \hline
    Striatum & 0.1326 & 0.1053 & 0.1133 & {\bf 0.0904} \\ \hline
    MRI & 0.0758 & 0.0642 & 0.0767 & {\bf 0.0545} \\ \hline   
    Natural & 0.1448 & 0.1389 & 0.1494 & {\bf 0.1240} \\
    \hline      
  \end{tabular}\\
  \caption{Variability of results by different AL strategies. 80\% of the scores are lying within the indicated interval. Feature Uncertainty is always more variable that Combined Uncertainty, batch selection is always less variable that single-instance selection. The best result is highlighted in bold.}
  \label{tab:06-variance} 
  \end{table}
\end{center}

In Fig.~\ref{fig:emresults},  we plot the  performance of all the  approaches we
consider in  terms of the  VOC~\cite{Pascal-voc-2010} score, a  commonly used measure  for this
kind of  application, as a  function of  the annotation effort.   The horizontal
line  at the  top depicts  the  VOC scores  obtained  by using  the {\it  whole}
training set,  which comprises 276130  and 325880 supervoxels, for  the striatum
and the hippocampus respectively.  \FU{} provides  a boost over \RS{}, and \CU{}
yields  a larger  one.  In  both  cases, a  further improvement  is obtained  by
introducing the batch-mode  geometry query selection of \PFU{}  and \PCU{}, with
the latter  coming on top. Recall  that these numbers are  averages over
  many  runs.    In  Table~\ref{tab:06-variance},  we  give   the  corresponding
  variances. Note that  both using the Geometric Uncertainty  and the batch-mode
  tend to reduce  them, thus making the process more  predictable. Note also the
  100  number we  use  very much  smaller  than the  total  number of  available
  samples.

Somewhat surprisingly, in the hippocampus case, the classifier performance given
only 100  training data points  is {\it higher} that  the one obtained  by using
{\it all} the training  data. In fact, this phenomenon has  been reported in the
AL  literature~\cite{Schohn00}  and  suggests  that  a  well  chosen  subset  of
datapoints can produce better generalisation performance than the complete set.

\subsection{Results on MRI data}
\label{sec:MRI}

Here  we  consider multimodal  brain  tumor  segmentation  in MRI  brain  scans.
Segmentation quality depends critically on the  amount of training data and only
highly-trained experts can provide it. T1,  T2, FLAIR, and post-Gadolinium T1 MR
images  are  available  in  the  BRATS  dataset for  each  one  of  20  subjects
\cite{Menze14}.   We use  standard filters  such as  Gaussian, gradient  filter,
tensor, Laplacian of  Gaussian and Hessian with different  parameters to compute
the feature vectors we feed to our Boosted Trees.

\begin{figure*}
\begin{center}
\begin{tabular}{cc}
\includegraphics[width=0.4\linewidth]{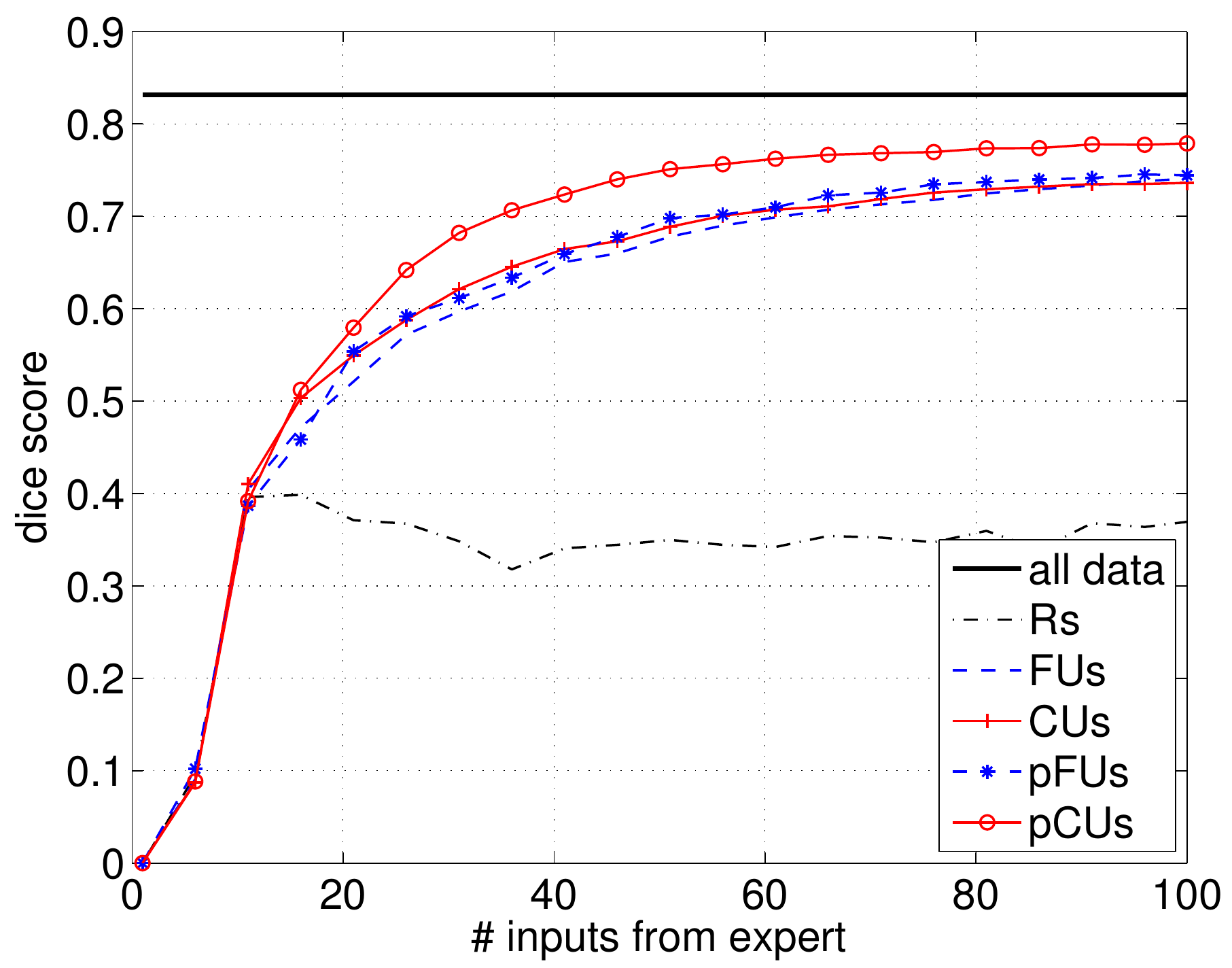} &
\includegraphics[width=0.4\linewidth]{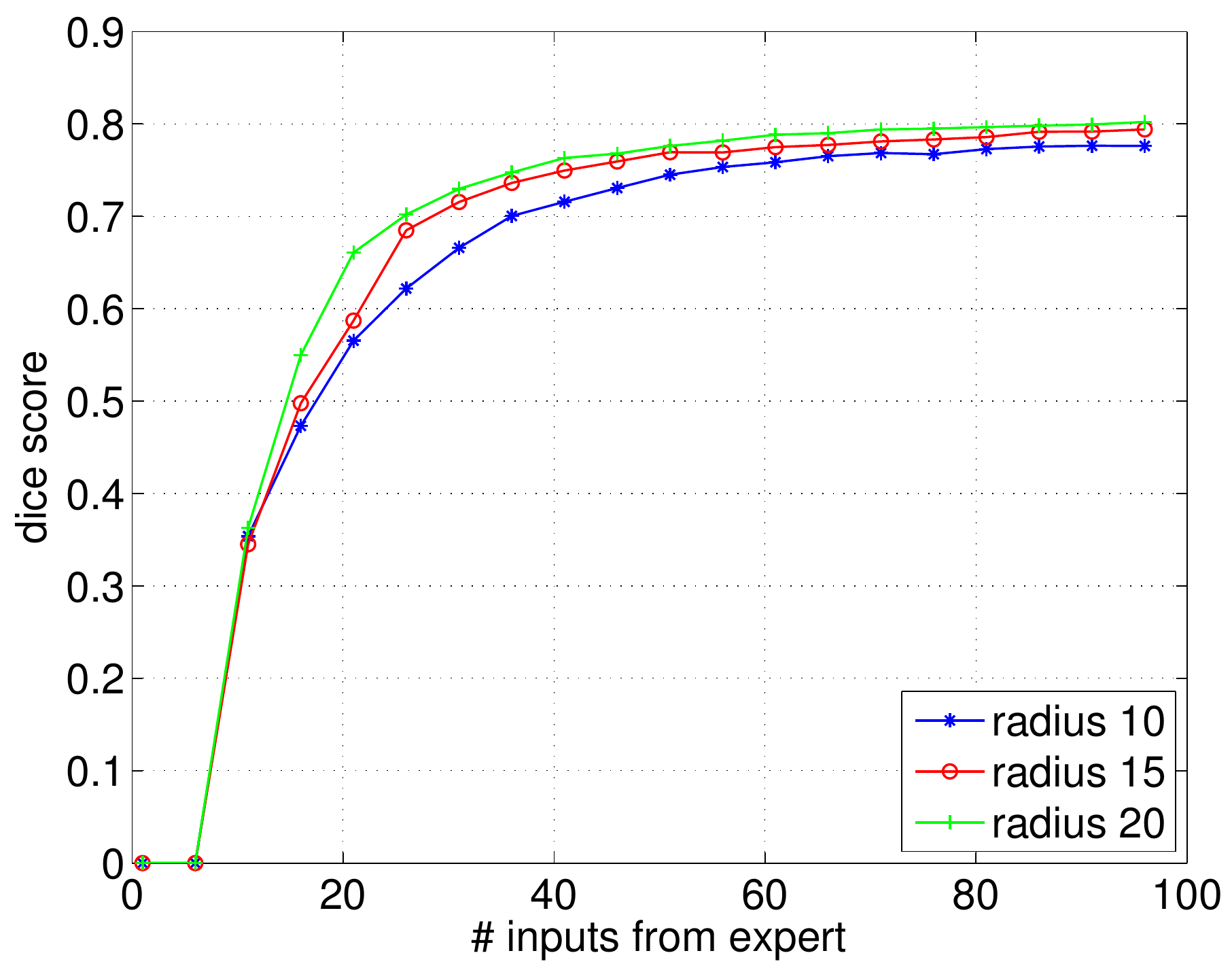}
\\
(a)&(b)
\end{tabular}
\caption{Comparison of various AL strategies for MRI data for tumor segmentation. Left: dice score for BRATS2012 dataset, right: $\PCU$ strategy with patches of different radius. }
\label{fig:mriresults}
\end{center}

\end{figure*}

In Fig.~\ref{fig:mriresults}, we  plot the performance of all  the approaches we
consider in terms  of the dice score~\cite{Gordillo13}, a  commonly used quality
measure for brain tumor segmentation, as a function of the annotation effort and
in Table~\ref{tab:06-variance}, we give the corresponding variances.  We observe
the same  pattern as in  Fig.~\ref{fig:emresults}, with \PCU{} again  doing best.  \comment{Shouldn't be  so much when I redraw it all on
  the same scale}

\comment{
First of all, we notice a huge gap
between on-line learning  with random sampling and any type  of Active Learning.
While random  selection of points  (\RS) improves the quality  measure very
slowly, uncertainty sampling \FU{} and \CU{} show much better
results, where including Geometric criteria (\CU{}) always dominating Feature Uncertainty (\FU{}).
Plane  selection doesn't  only make  the labelling  procedure easier,  but
also boosts the performance  and makes the benefit of  Geometric Uncertainty (\PCU{})
more visible.
} 

\begin{figure}[b]
\begin{center}
   \includegraphics[width=0.95\linewidth]{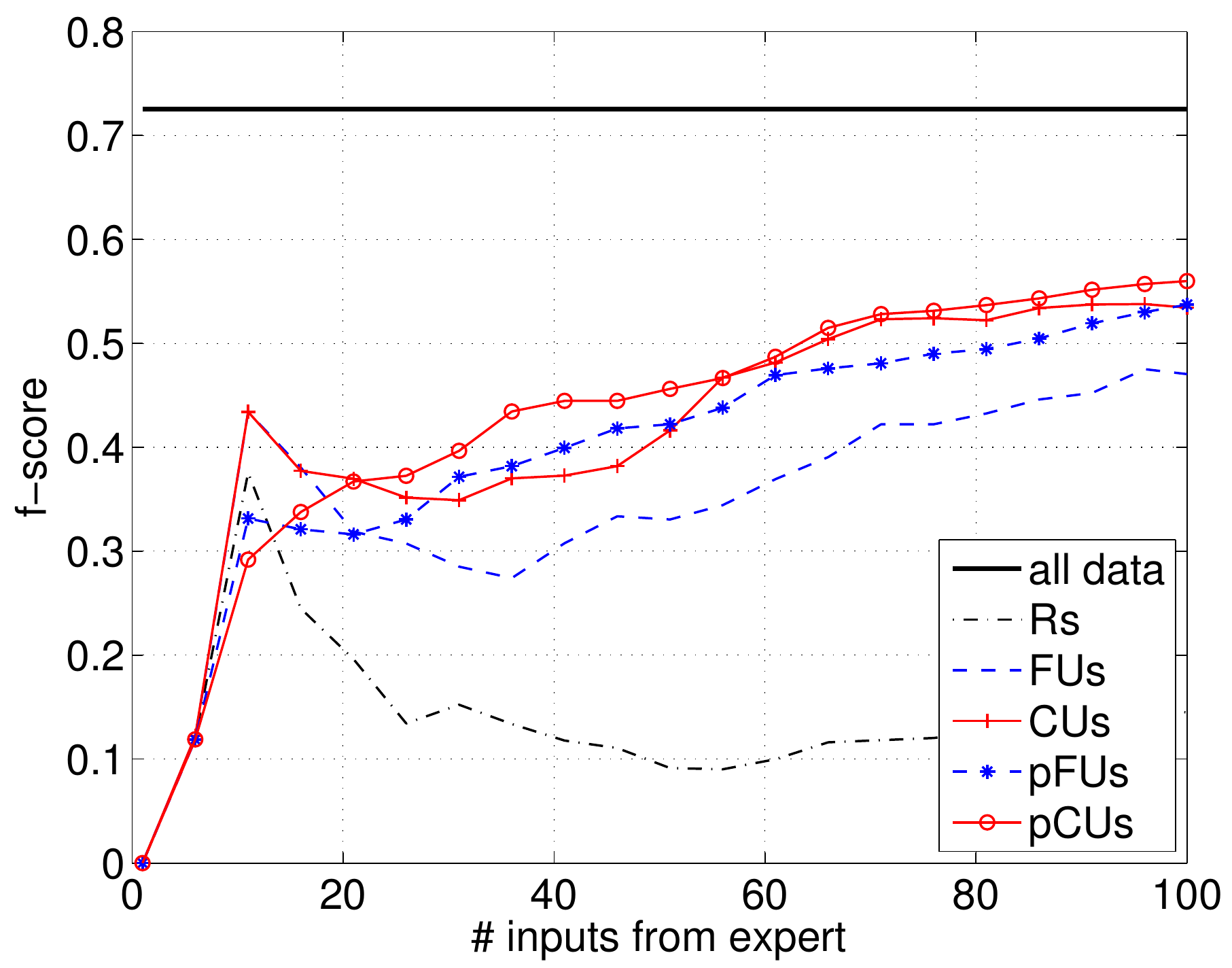}
\end{center}
   \caption{Comparison of various AL strategies for segmentation of natural images.}
   \label{fig:06-naturalresults}
\end{figure}


The patch radius parameter $r$ of Sec.~\ref{sec:BatchGeom} plays an important role in
plane selection. To  evaluate its  influence, we  recomputed our  \PCU{} results  50 times
using three different values for $r=10$,  $15$ and $20$.  The resulting plots are
shown in Fig.~\ref{fig:mriresults}. With  a larger radius, the learning-rate is
slightly higher  as could be  expected from since  more voxels are  labeled each
time. However, as the patches become larger,  it stops being clear that this can
be done  with only  two mouse  clicks and that  is why  we limited  ourselves to
radius sizes of 10 to 15.

\subsection{Natural Images}

Finally, we turn to natural 2D images and replace supervoxels by superpixels.
In this case, the plane selection of  \PFU{} and \PCU{} reduces to simple selection
of image patches  in the image.  In practice, we  simply select superpixels with
their 4 neighbors. 
Increasing this number would lead to higher learning rates in the same way as increasing the patch radius $r$, but we restrics it to a small value to ensure labelling can be done with 2 mouse clicks on average.
To compute image features, we
use Gaussian, Laplacian, Laplacian of Gaussian, Prewitt, Sobel filters to filter
intensity and color values, gather first-order statistics such as local standard
deviation, local range, gradient magnitude  and direction histograms, as well as
SIFT  features.

We    plot    our    results    on    the    Weizmann    horse    database    in
Fig.~\ref{fig:06-naturalresults}  and   give  the  corresponding   variances  in
Table~\ref{tab:06-variance}. The pattern is again  similar to the one observed in
Figs.~\ref{fig:emresults}  and~\ref{fig:mriresults}, with  the difference  between
\CU{} and \PCU{}  being smaller due to  the fact that 2D  batch-mode approach is
much less  sophisticated than  the 3D  one.  Note, however,  that the  first few
iterations are disastrous for all methods, however, plane-based methods are able to recover from it quite fast. 
\comment{ and that simple \FU{} does even worse
than  even simpler  \RS{}, which  is also  a phenomenon  that has  been reported
previously~\cite{Baldridge09}. To check that this  is not caused by our approach
to choosing  a threshold for the  purpose of computing the  entropy as described
in Sec.~\ref{sec:setup},  we tried  an  additional experiment  by  having a  user
provide an  optimal threshold value after  every iteration but this  did produce
any significant change for $\FU$. }


\section{Conclusion}

In  this paper  we introduced  an approach  to exploiting  the geometric  priors
inherent  to  images  to  increase  the effectiveness  of  Active  Learning  for
segmentation purposes.  For  2D images, it relies on an  approach to Uncertainty
Sampling  that accounts  not only  for the  uncertainty of  the prediction  at a
specific location but also in its neighborhood.  For 3D image stacks, it adds to
this  the  ability to  automatically  select  a  planar  patch in  which  manual
annotation is easy to do.

We have formulated our algorithms in terms of background/foreground segmentation
but the  entropy functions that we  use to express our  uncertainties can handle
multiple classes with little change to  the overall approach. In future work, we
will therefore extend our approach to more general segmentation problems.


\section*{Acknowledgements}

We would like to thank Carlos Becker and Lucas Maystre for useful and inspiring discussions, Agata Mosinska and R\'{o}ger Berm\'{u}dez-Chac\'{o}n for their proofreading and comments on the text.

\clearpage
{\small
\bibliographystyle{ieee}
\bibliography{string,vision,learning,biomed,optim,misc}

\begin{thebibliography}{10}\itemsep=-1pt

\bibitem{Achanta12}
R.~Achanta, A.~Shaji, K.~Smith, A.~Lucchi, P.~Fua, and S.~Suesstrunk.
\newblock {{SLIC} Superpixels Compared to State-Of-The-Art Superpixel Methods}.
\newblock {\em IEEE Transactions on Pattern Analysis and Machine Intelligence},
  34(11):2274--2282, November 2012.

\bibitem{Altaie14}
A.~Al-Taie, H.~H. K., and L.~Linsen.
\newblock {Uncertainty Estimation and Visualization in Probabilistic
  Segmentation}.
\newblock 2014.

\bibitem{Andres08}
B.~Andres, U.~Koethe, M.~Helmstaedter, W.~Denk, and F.~Hamprecht.
\newblock {Segmentation of {SBFSEM} Volume Data of Neural Tissue by
  Hierarchical Classification}.
\newblock In {\em DAGM Symposium on Pattern Recognition}, pages 142--152, 2008.

\bibitem{Becker13b}
C.~Becker, R.~Rigamonti, V.~Lepetit, and P.~Fua.
\newblock {Supervised Feature Learning for Curvilinear Structure Segmentation}.
\newblock In {\em Conference on Medical Image Computing and Computer Assisted
  Intervention}, September 2013.

\bibitem{Elhamifar13}
E.~Elhamifar, G.~Sapiro, A.~Yang, and S.~S. Sasrty.
\newblock {A Convex Optimization Framework for Active Learning}.
\newblock In {\em International Conference on Computer Vision}, 2013.

\bibitem{Pascal-voc-2010}
M.~Everingham, C.~W. L.~{Van~Gool}~and, J.~Winn, and A.~Zisserman.
\newblock {The Pascal Visual Object Classes Challenge~{(VOC2010)} Results},
  2010.

\bibitem{GiladBachrach05}
R.~Gilad-Bachrach, A.~Navot, and N.~Tishby.
\newblock {Query By Committee Made Real}.
\newblock In {\em Advances in Neural Information Processing Systems}, 2005.

\bibitem{Gordillo13}
N.~Gordillo, E.~Montseny, and P.~Sobrevilla.
\newblock {State of the Art Survey on MRI Brain Tumor Segmentation}.
\newblock {\em Magnetic Resonance in Medicine}, 2013.

\bibitem{Grady06}
L.~Grady.
\newblock {Random Walks for Image Segmentation}.
\newblock {\em IEEE Transactions on Pattern Analysis and Machine Intelligence},
  28(11):1768--1783, 2006.

\bibitem{Hastie01}
T.~Hastie, R.~Tibshirani, and J.~Friedman.
\newblock {\em {The Elements of Statistical Learning}}.
\newblock Springer, 2001.

\bibitem{Hoi06}
S.~C.~H. Hoi, R.~Jin, J.~Zhu, and M.~R. Lyu.
\newblock {Batch Mode Active Learning and its Application to Medical Image
  Classification}.
\newblock In {\em International Conference on Machine Learning}, 2006.

\bibitem{Iglesias11}
J.~Iglesias, E.~Konukoglu, A.~Montillo, Z.~Tu, and A.~Criminisi.
\newblock {Combining Generative and Discriminative Models for Semantic
  Segmentation}.
\newblock In {\em Information Processing in Medical Imaging}, 2011.

\bibitem{Joshi09}
A.~Joshi, F.~Porikli, and N.~Papanikolopoulos.
\newblock {Multi-Class Active Learning for Image Classification}.
\newblock In {\em Conference on Computer Vision and Pattern Recognition}, 2009.

\bibitem{Kapoor07}
A.~Kapoor, K.~Grauman, R.~Urtasun, and T.~Darrell.
\newblock {Active Learning with Gaussian Processes for Object Categorization}.
\newblock In {\em International Conference on Computer Vision}, 2007.

\bibitem{Lewis94}
D.~Lewis and W.~Gale.
\newblock {A Sequential Algorithm for Training Text Classifiers}.
\newblock 1994.

\bibitem{Li11}
Q.~Li, Z.~Deng, Y.~Zhang, X.~Zhou, U.~V. Nagerl, and S.~T.~C. Wong.
\newblock {A Global Spatial Similarity Optimization Scheme to Track Large
  Numbers of Dendritic Spines in Time-Lapse Confocal Microscopy}.
\newblock {\em IEEE Transactions on Medical Imaging}, 30(3):632--641, 2011.

\bibitem{Lin14}
T.-Y. Lin, M.~Maire, S.~Belongie, J.~Hays, P.~Perona, D.~Ramanan,
  P.~Doll{\'a}r, and C.~Zitnick.
\newblock {Microsoft COCO: Common objects in context}.
\newblock In {\em European Conference on Computer Vision}, pages 740--755,
  2014.

\bibitem{Long13}
C.~Long, G.~Hua, and A.~Kapoor.
\newblock {Active Visual Recognition with Expertise Estimation in
  Crowdsourcing}.
\newblock In {\em International Conference on Computer Vision}, 2013.

\bibitem{Lovasz93}
L.~Lov{\'a}sz.
\newblock {Random Walks on Graphs: A Survey}.
\newblock {\em {Combinatorics, Paul Erdos is Eighty}}, 1993.

\bibitem{Lucchi11b}
A.~Lucchi, K.~Smith, R.~Achanta, G.~Knott, and P.~Fua.
\newblock {Supervoxel-Based Segmentation of Mitochondria in EM Image Stacks
  with Learned Shape Features}.
\newblock {\em IEEE Transactions on Medical Imaging}, 31(2):474--486, February
  2012.

\bibitem{Maiora12}
J.~Maiora and M.~G. {n}a.
\newblock {Abdominal CTA Image Analysis through Active Learning and Decision
  Random Forests: Aplication to AAA Segmentation}.
\newblock In {\em IJCNN}, 2012.

\bibitem{Menze14}
B.~Menza, A.~Jacas, et~al.
\newblock {The Multimodal Brain Tumor Image Segmentation Benchmark (BRATS)}.
\newblock {\em IEEE Transactions on Medical Imaging}, 2014.

\bibitem{Olabarriaga01}
S.~D. Olabarriaga and A.~W.~M. Smeulders.
\newblock {Interaction in the Segmentation of Medical Images : A Survey}.
\newblock {\em Medical Image Analysis}, 2001.

\bibitem{Olsson09}
F.~Olsson.
\newblock {A Literature Survey of Active Machine Learning in the Context of
  Natural Language Processing}.
\newblock {\em Swedish Institute of Computer Science}, 2009.

\bibitem{Schmid10}
B.~Schmid, J.~Schindelin, A.~Cardona, M.~Longair, and M.~Heisenberg.
\newblock {A High-Level 3D Visualization {API} for {J}ava and {ImageJ}}.
\newblock {\em BMC Bioinformatics}, 11:274, 2010.

\bibitem{Schohn00}
G.~Schohn and D.~Cohn.
\newblock Less is more: Active learning with support vector machines.
\newblock In {\em ICML}, 2000.

\bibitem{Settles10}
B.~Settles.
\newblock {Active Learning Literature Survey}.
\newblock Computer Sciences Technical Report 1648, University of
  Wisconsin--Madison, 2010.

\bibitem{Settles11}
B.~Settles.
\newblock {From Theories to Queries : Active Learning in Practice}.
\newblock {\em Active Learning and Experimental Design}, 2011.

\bibitem{Settles08}
B.~Settles, M.~Craven, and S.~Ray.
\newblock {Multiple-Instance Active Learning}.
\newblock In {\em Advances in Neural Information Processing Systems}, 2008.

\bibitem{Sznitman13a}
R.~Sznitman, C.~Becker, F.~Fleuret, and P.~Fua.
\newblock {Fast Object Detection with Entropy-Driven Evaluation}.
\newblock In {\em Conference on Computer Vision and Pattern Recognition}, pages
  3270--3277, 2013.

\bibitem{Sznitman10}
R.~Sznitman and B.~Jedynak.
\newblock {Active Testing for Face Detection and Localization}.
\newblock {\em IEEE Transactions on Pattern Analysis and Machine Intelligence},
  32(10):1914--1920, June 2010.

\bibitem{Tong02}
S.~Tong and D.~Koller.
\newblock {Support Vector Machine Active Learning with Applications to Text
  Classification}.
\newblock {\em Machine Learning}, 2002.

\bibitem{Top11b}
A.~Top, G.~Hamarneh, and R.~Abugharbieh.
\newblock {Active learning for interactive 3D image segmentation}.
\newblock {\em Conference on Medical Image Computing and Computer Assisted
  Intervention}, 2011.

\bibitem{Vezhnevets12}
A.~Vezhnevets, J.~Buhmann, and V.~Ferrari.
\newblock {Active Learning for Semantic Segmentation with Expected Change}.
\newblock In {\em Conference on Computer Vision and Pattern Recognition}, 2012.

\end{thebibliography}
}

\end{document}